\pdfoutput=1
\documentclass[11pt]{article}

\usepackage[final]{acl}

\usepackage{times}
\usepackage{latexsym}

\usepackage[T1]{fontenc}
\usepackage[utf8]{inputenc}

\usepackage{microtype}

\usepackage{inconsolata}

\usepackage{graphicx}

\usepackage{booktabs}
\usepackage{multirow}
\usepackage{amsmath}
\usepackage{amssymb}
\usepackage{fontawesome}

\title{Beyond Static Testbeds: An Interaction-Centric Agent Simulation Platform for Dynamic Recommender Systems}

\author{
   Song Jin\textsuperscript{1}\thanks{\ \  Equal contribution to this work.}, Juntian Zhang\textsuperscript{1}\footnotemark[1], Yuhan Liu\textsuperscript{1}\thanks{\ \ Corresponding authors.}, \\
   \textbf{Xun Zhang}\textsuperscript{2}, \textbf{Yufei Zhang}\textsuperscript{2},\textbf{Guojun Yin}\textsuperscript{2}, \textbf{Fei Jiang}\textsuperscript{2}, \textbf{Wei Lin}\textsuperscript{2},
  \textbf{Rui Yan}\textsuperscript{1,3}\footnotemark[2] \\
  \textsuperscript{1}Gaoling School of Artificial Intelligence, Renmin University of China, \\ \textsuperscript{2}Meituan, \textsuperscript{3}Wuhan University\\
  \texttt{\{jinsong8, zhangjuntian\}@ruc.edu.cn}
}

\begin{document}
\maketitle
\begin{abstract}

Evaluating and iterating upon recommender systems is crucial, yet traditional A/B testing is resource-intensive, and offline methods struggle with dynamic user-platform interactions. While agent-based simulation is promising, existing platforms often lack a mechanism for user actions to dynamically reshape the environment. To bridge this gap, we introduce \textbf{RecInter}, a novel agent-based simulation platform for recommender systems featuring a robust interaction mechanism. In \textbf{RecInter} platform, simulated user actions (e.g., likes, reviews, purchases) dynamically update item attributes in real-time, and introduced \textit{Merchant Agents} can reply, fostering a more realistic and evolving ecosystem. High-fidelity simulation is ensured through\textit{ Multidimensional User Profiling module}, \textit{Advanced Agent Architecture}, and LLM fine-tuned on Chain-of-Thought (CoT) enriched interaction data. Our platform achieves significantly improved simulation credibility and successfully replicates emergent phenomena like \textit{Brand Loyalty} and the \textit{Matthew Effect}. Experiments demonstrate that this interaction mechanism is pivotal for simulating realistic system evolution, establishing our platform as a credible testbed for recommender systems research: \faGithub~\href{https://github.com/jinsong8/RecInter}{RecInter}.

\end{abstract}

\section{Introduction}
Recommender systems are integral to modern digital platforms, shaping user experiences and driving engagement across diverse domains. The ability to effectively evaluate and iterate upon these systems is paramount for their continued improvement and success~\cite{ricci2010introduction}. However, traditional online A/B testing, while considered the gold standard, is often resource-intensive, time-consuming, and can carry risks associated with deploying untested algorithms to real users. Offline evaluation methods, conversely, struggle to capture the dynamic, interactive nature of user-platform engagement. Agent-based simulation has emerged as a powerful and cost-effective paradigm to bridge this gap,
% due to the strong reasoning capabilities of LLMs~\cite{zhang2025weaving}, 
offering a controlled environment to test new algorithms, understand user behavior, and explore complex system dynamics such as filter bubbles, information cocoons, and user conformity behaviors~\cite{zhang2024generative, wang2025user}. 

Early recommendation simulation, like Virtual Taobao~\cite{Virtual-Taobao} and RecSim~\cite{RecSim}, primarily 
% served as data generators or testbeds for reinforcement learning, often 
rely on rule-based user models that, while useful, limited their behavioral realism and adaptability. The recent advent of Large Language Models (LLMs) has catalyzed a new wave of sophistication in agent-based modeling~\cite{Wang2023survey, xi2025rise}. 
% LLM-powered agents, as demonstrated by frameworks like Generative Agent~\cite{Generative_Agent}, can exhibit more nuanced, human-like behaviors. 
In the recommendation domain, RecAgent~\cite{wang2025user} pioneered the use of LLM agents to simulate diverse user activities on a platform. Agent4Rec~\cite{zhang2024generative} and SimUSER~\cite{bougie2025simuser} enhance the reliability of user behavior simulation in recommender systems by incorporating enriched user profiling and specially designed memory mechanisms. However, a critical aspect often underrepresented in existing recommendation simulations is a deeply integrated interaction mechanism. Real-world platforms are not static; item attributes (e.g., popularity, average ratings, review sentiment) evolve dynamically based on continuous user feedback and even merchant interventions.  Research~\cite{lee2021product, cui2012effect, li2019effect} has already demonstrated the significant impact of these dynamic item attributes on user decision-making.
This reciprocal relationship, where user actions reshape the environment and, in turn, influence future user decisions, 
is crucial for realistic simulation.

To address this gap, we introduce \textbf{RecInter}, a novel agent-based simulation platform for recommender systems incorporating a interaction mechanism. Our platform allows simulated user actions such as liking, reviewing, or purchasing items to dynamically update item attributes in real time. Furthermore, we introduce \textit{Merchant Agents} capable of replying to users, further enriching the dynamic nature of the simulated ecosystem. To achieve high-fidelity simulations, we integrate three components. Firstly, a \textit{Multidimensional User Profiling} module to construct detailed user profiles from historical data. Secondly, our platform's User Agents are engineered with \textit{sophisticated memory systems} (encompassing both perceptual and cognitive faculties) and advanced action selection mechanisms to more accurately emulate human decision-making processes. Thirdly, a \textit{Behavior Simulation Training} pipeline is employed to fine-tune the LLM-based agents using high-quality, CoT enriched interaction data. 

Our contributions are threefold:

$\bullet$ We develop a realistic recommendation simulation platform \textbf{RecInter} featuring a novel interaction mechanism where user feedback and merchant replies dynamically alter item attributes, fostering a more lifelike and evolving environment.
    
$\bullet$ Through Multidimensional User Profiling, Advanced Agent Architecture and  Behavior Simulation Training, we achieve a higher authentication of simulated user behaviors, significantly surpassing previous methods.
    % This is validated by extensive metric-based, LLM-based, and macro-level evaluations.
    
$\bullet$ We explored and validated the crucial role of the interaction mechanism in modeling realistic system using \textbf{RecInter}. \textbf{RecInter} successfully reproduced the Brand Loyalty and Matthew Effect phenomena, demonstrating its credibility as a reliable testbed for recommendation systems research.

\section{Related Work}

\subsection{LLM-based Agents}

Large Language Model (LLM) agents are autonomous computational entities that perceive, decide, and act within their environment~\cite{xi2025rise}. 
% Early works on intelligent agents have laid foundational concepts for decision-making and collaboration \cite{jennings1998roadmap, franklin1996agent, panait2005cooperative}. 
With the emergence of LLMs, agent-based systems have gained renewed attention~\cite{Wang2023survey}. The Generative Agent framework~\cite{Generative_Agent} introduced agents with memory, planning, and reflection, simulating human cognition. Recent efforts divide LLM agents into task-oriented and simulation-oriented categories~\cite{xi2025rise}.

Task-oriented agents are designed to accomplish specific goals defined by users. For example, Voyager~\cite{Voyager} enables LLM-driven navigation in Minecraft, while ChatDev~\cite{ChatDev} and AutoGen~\cite{AutoGen} build collaborative multi-agent systems for software development.
% , incorporating advanced role assignments and communication. 
Simulation-oriented agents, on the other hand, focus on modeling human-like behaviors and social dynamics. The SANDBOX framework~\cite{Socially_alignment} explores social issues through multi-agent interaction to support LLM alignment with ethical norms, while WereWolf and AgentSims~\cite{WereWolf, AgentSims} use game-like environments to examine complex group dynamics. The FPS~\cite{liu2024skepticism} and FUSE~\cite{liu2024tiny} respectively explore the use of LLMs-based agent for simulating the propagation and the evolution of fake news. Our framework provides a more accurate simulation of user reviews and merchant responses in recommendation scenarios.

\subsection{Recommendation Simulation}

Recommendation simulators have emerged as a cornerstone in recommender systems research~\cite{reinforcement_survey_simulator, top_n_simulator, mindsim_simulator,liu2025truth}, offering a cost-effective alternative to online testing and addressing persistent challenges such as serendipitous discovery and filter bubbles~\cite{debias_simulator, adver_user_model_simulator}. Early simulators primarily served as data sources for reinforcement learning applications. Notable examples include Virtual Taobao~\cite{Virtual-Taobao}, which simulates e-commerce user behaviors, and RecSim~\cite{RecSim}, providing toolkits for sequential recommendation simulations.
% , and RecoGym~\cite{RecoGym}, integrating traditional algorithms with reinforcement learning frameworks. 
% MINDSim~\cite{mindsim_simulator} extended this approach to news recommendation scenarios. 
However, these conventional simulators often relied on simplistic rules, limiting their flexibility and validity. The recent advent of LLM-powered agents has shown remarkable potential in approximating human-like intelligence~\cite{Wang2023survey}, opening new avenues for more sophisticated recommendation simulators. A notable example is RecAgent~\cite{wang2025user}, which pioneered the development of a recommendation platform integrating diverse user behaviors.
% such as movie-watching, chatting, posting, and searching. 
Agent4Rec~\cite{zhang2024generative} proposes an agent system composed of LLMs to simulate recommendation systems, and SimUSER~\cite{bougie2025simuser} is an agent framework that simulates human-like behavior to evaluate recommender algorithms, using self-consistent personas and memory modules.
% % for more realistic assessments.
Building on these advancements, our research explores how user feedback and merchant replies dynamically influence item attributes, enabling more realistic recommendation simulations.

\section{Methodology}
\begin{figure*}
 \centering
\includegraphics[width=0.93\textwidth]{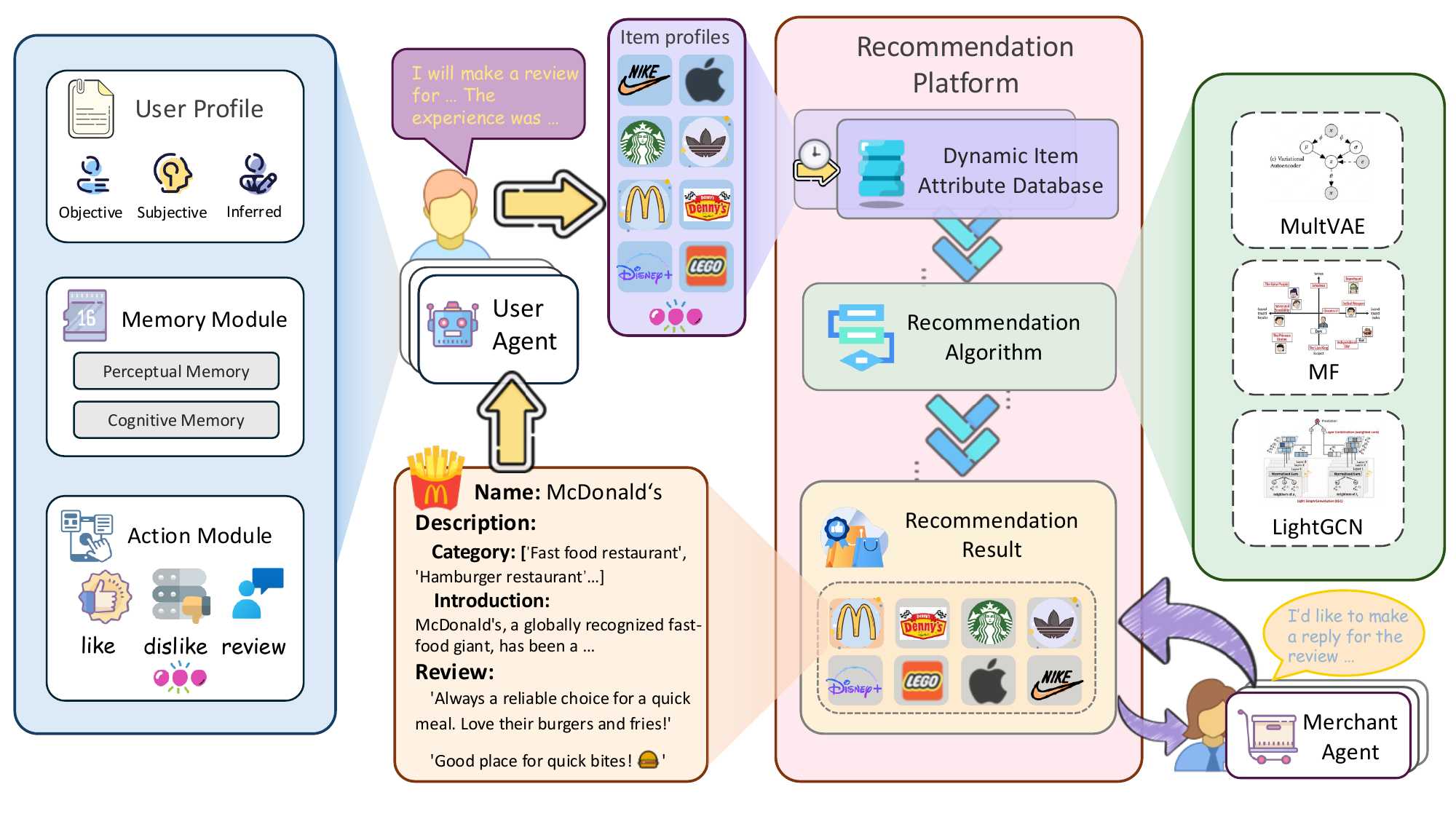}
\caption{The overall framework of \textbf{RecInter}. The User Agent, equipped with user profile, memory module and action module, interacts with the platform by taking actions that can modify the attributes of items. In response, the platform guided by recommendation algorithm returns updated items to the user, thereby completing the interaction loop. Similarly, the Merchant Agent is also capable of  participating in this dynamic process.}
\vspace{-0.4cm}
\label{fig:data_collection_construction}
\end{figure*}

Our simulation platform, \textbf{RecInter} (as illustrated in Figure~\ref{fig:data_collection_construction}), is designed to emulate a realistic recommendation scenario. To achieve this objective, we focus on two key aspects: (1) enhancing the credibility of user simulation, and (2) constructing a interactive recommendation platform environment. To improve user simulation accuracy, we introduce modules including Multidimensional User Profiling, an Advanced User Agent Architecture, and a Behavior Simulation Training pipeline. Additionally, we build an interactive recommendation platform environment that incorporates dynamic updates, merchant reply, and recommendation algorithm.

\subsection{Problem Formulation} 

Let $\mathcal{U}$ denote the set of users and $\mathcal{I}$ represent the item set. For each user $u \in \mathcal{U}$, we first extract user profiles from their historical interaction sequences $\mathcal{H}_u = \{(i_1, r_1, c_1), (i_2, r_2, c_2), ..., (i_{N_u}, r_{N_u}, c_{N_u})\}$, where $i_j \in \mathcal{I}$ represents an interacted item with its rich contextual information, $r_j \in \{1, 2, 3, 4, 5\}$ denotes the rating provided by the user, and $c_j$ denotes the textual review provided by the user for item $i_j$. We construct a user profile pool 
$\mathcal{P}=\{P(u)|u \in U\}$, where each profile $P(u)$ encodes information extracted from the user's historical interactions. This pool forms the basis for instantiating simulated user agents. \textbf{RecInter} operates for $T$ time steps. At each time step $t$, the platform recommends a set of items $\mathcal{R}_t \subset \mathcal{I}$ to the simulated user agent, who then provides feedback $\mathcal{F}_t$ based on their preferences. This feedback subsequently updates the attributes $A$ of items on the platform. Our objective is to minimize the behavioral discrepancy $\mathcal{D}(B_{real}, B_{sim})$ between the simulated user agents and real users, thereby creating a realistic simulation environment for recommender systems research.

\subsection{Multidimensional User Profiling}
\begin{figure}
\centering
\includegraphics[width=0.7\linewidth]{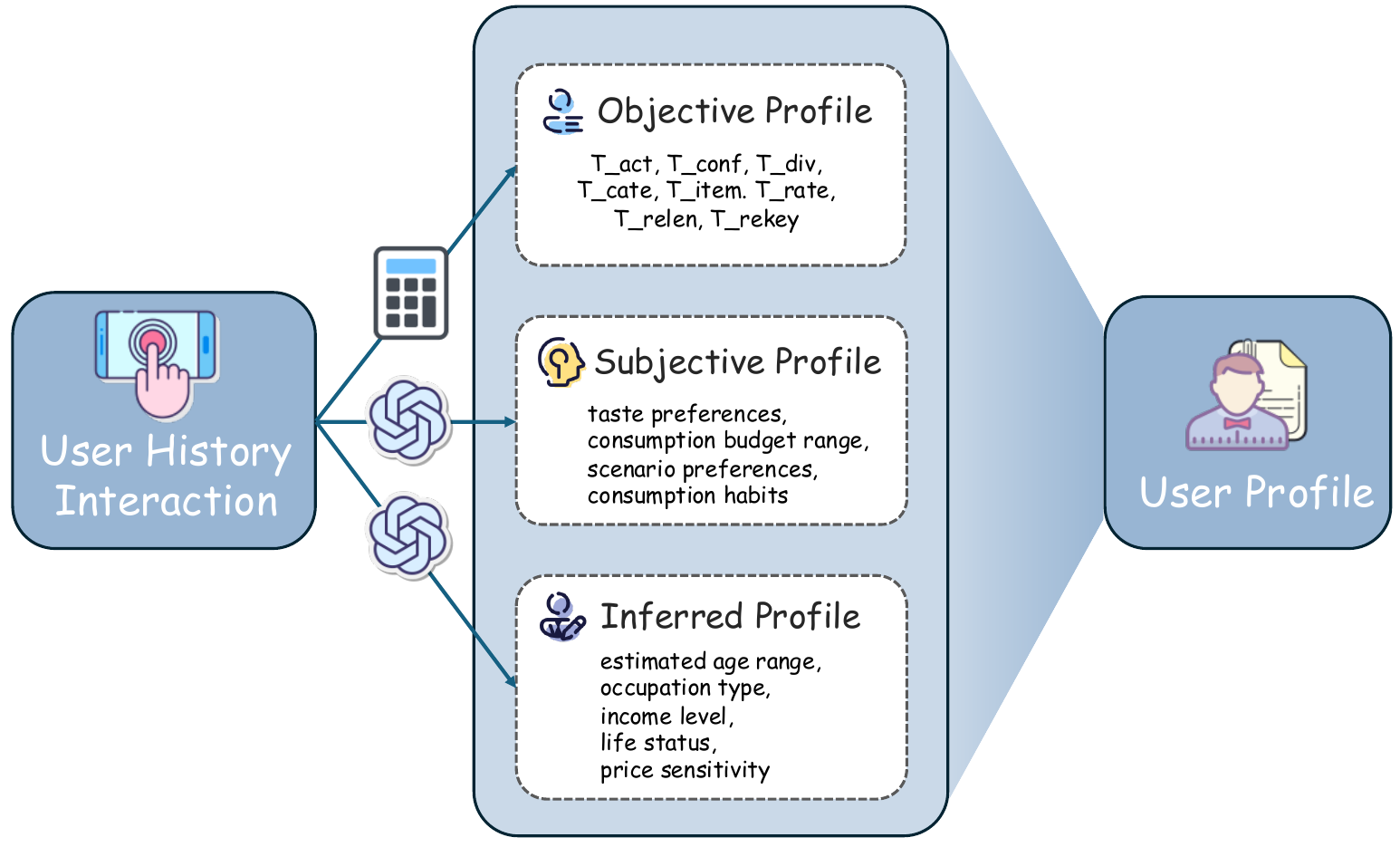}
\caption{Multidimensional User Profiling framework.}
\vspace{-0.4cm}
\label{fig:framework}
\end{figure}

As presented in the Figure~\ref{fig:framework}, Multidimensional User Profiling involves constructing the user's objective, subjective, and inferred profile.

\noindent \textbf{Objective Profile}
% Both \cite{zhang2024generative} and \cite{bougie2025simuser} have demonstrated that having accurate and realistic user profiles is crucial for effective user simulation in recommender systems. In particular, \cite{zhang2024generative} highlights that certain statistical metrics constitute essential components of user profiles. They define activity $T_{\text{act}}$ to assess user enagement, conformity $T_{\text{conf}}$ to indicate the deviation between user and mainstream opinions, $T_{\text{div}}$ to reflect user diversified preferences. Building upon the metrics, we also introduce several new indicators to further enhance the realism of user profiles. We conducted a systematic analysis of product categories within user interaction history 
Building on the statistical metrics $T_{\text{act}}$, $T_{\text{conf}}$, and $T_{\text{cons}}$ proposed by \citet{zhang2024generative}, we further introduce a set of novel indicators aimed at enhancing the realism of user objective profile. We conducted a systematic analysis of product categories within user interaction history $\mathcal{H}_u$ and identified the most top-k frequently interacted item categories as $T_{\text{cate}}= \{c_1,c_2,...,c_{k}\}$, where $k$ is set to 30. We also calculated the top-k most frequently interacted items as $T_{\text{item}} = \{i_1, i_2, ..., i_{k}\}$, where $k$ is set to 10. We use $T_{\text{rate}} = \frac{1}{k} \sum_{i=1}^{k} \text{r}_i$ to calculate the user's historical average rate score. In our designed \textbf{RecInter}, users are able to leave reviews for items. Therefore, it is necessary to additionally consider the characteristics of user reviews. We define the probability that a user leaves a review as $T_{repr} = \frac{1}{k} \sum_{i=1}^{k} \mathbb{I}(r_i \neq \emptyset)$. To represent the average length of a user's historical reviews, we use $T_{\text{relen}} = \frac{1}{k} \sum_{i=1}^{k} \text{len}(r_i)$, where $\text{len}(r_i)$ denotes the length of the $i$-th review. Additionally, we explore the user's review style by extracting the top-N most frequent keywords from their reviews using TF-IDF~\cite{salton1988term}. We denote the set of these keywords as $T_{\text{rekey}} = \{w_1, w_2, ..., w_{N}\}$, where $N$ is set to 20.

\noindent \textbf{Subjective Profile}
While objective profile constitute a crucial part of user profile, relying solely on statistical indicators often fails to capture the more nuanced aspects of user preferences. To address this, we leverage LLM to derive subjective user profile. Specifically, for each user, we randomly sample 60 items from their historical interactions $\mathcal{H}_u$ and apply an LLM GPT-4o to perform information augmentation, generating more detailed descriptions for these items. The items’ basic information $A$, augmented content $\bar{A}$, and user ratings $r$ are provided as inputs to the LLM to facilitate the construction of subjective user profiles.  This approach enables the model to summarize key aspects of the subjective profile, including taste preferences, consumption budget range, scenario preferences, and consumption habits. 
The prompt used and one profile case are provided in Appendix~\ref{subjective_prompt} and Appendix~\ref{subjective_profile_case} respectively.

\noindent \textbf{Inferred Profile}
User reviews have been shown to indirectly reflect personal profile~\cite{sachdeva2020useful, srifi2020recommender}. 
% For example, a complaint about waiting time may indicate high time sensitivity, while a mention of satisfaction shared with a spouse may imply marital status and age range. 
Despite this, prior user simulation in recommender systems have largely ignored the potential of review data. To address this, we leverage LLM with carefully designed prompt to infer user profile from reviews. The input includes 60 items with each item's basic attributes $A$, augmented content $\bar{A}$, user ratings $r$, and review texts $c$. To reduce hallucinations, the model is instructed to output ``unknow'' when inference is uncertain. This process yields inferred profile elements such as estimated age range, occupation type, income level, life status, price sensitivity, quality consciousness, service preferences, points of concern, and review language style. The prompt used and one inferred profile case are provided in Appendix~\ref{infer_prompt} and Appendix~\ref{inferred_profile_case} respectively.

\subsection{Advanced User Agent Architecture}
\subsubsection{Memory Module}
% Recent studies \cite{park2023generative, liu2024skepticism, liu2024lmagent} emphasize the importance of memory modules in simulated agents, using short-term memory, long-term memory and reflection to improve performance. 
Recommender systems face unique challenges, including vast number of items, shifting user interests, and the influence of users' cognitive states. To address this, we introduce perceptual memory and cognitive memory.
% , tailored to capture user interactions and underlying psychological signals for improved user modeling.

\noindent \textbf{Perceptual Memory} 
Perceptual memory stores an agent's historical interactions and serves as the agent’s past experiences. To enable faster and more accurate retrieval while conserving prompt space, we perform a simplification process that retains only the most essential information from each interaction. Specifically, each perceptual memory is represented as $m_t^p = (l_t, a_t,t)$, where $l_t = \{i_1, i_2, ..., i_n\}$ denotes the list of recommended items received by the agent at time step $t$, $a_t$ represents the agent's action at that time.

\noindent \textbf{Cognitive Memory}
In \textbf{RecInter}, agents engage in internal reasoning and judgment before taking actions, analogous to how individuals think before making a purchase. This internal cognitive process influences both current and future decisions. For instance, when presented with two chip flavors, a user may choose one now while planning to try the other later. To capture this process, we define cognitive memory at time step t as $m_t^c = (\bar{l_t}, s_t, a_t, t)$, where $\bar{l_t}$ is the list of recommended items with their attributes, $s_t$ is the agent’s thought process, $a_t$ is the action taken, and $t$ is the time step.

At time step t, the agent retrieves relevant information from its perceptual memory $M^p = [m_1^p, m_2^p, ..., m_{N_p}^p]$ and cognitive memory $M^c = [m_1^c, m_2^c, ..., m_{N_c}^c]$ before making action. Following \citet{park2023generative}, each memory is assigned an importance score:
\[
score_i = \alpha e^{-\gamma \cdot \nabla t} + \beta \text{sim}(m_i, \mathcal{R}_t)
\]
where $\nabla t$ is the time elapsed since the memory was formed, $\text{sim}(m_i, \mathcal{R}_t)$ denotes the similarity between memory $m_i$ and the current reasoning context $\mathcal{R}_t$, and $\alpha, \beta, \gamma$
are tunable parameters. In addition to retrieving the top-scoring memories, we impose adaptive thresholds $\theta_p$ and $\theta_c$ to dynamically adjust the maximum number of retrieved memory for each memory type. The specific hyperparameter is included in Appendix~\ref{app:hyperparameter}.

\subsubsection{Action Module}
% \cite{yang2024oasis} demonstrated that a "think-then-act" approach,similar to chain-of-thought reasoning, enables simulated users to behave more similarly to real users. While their focus was on user simulation within social networks, we argue that such a paradigm is more vital in the context of recommender systems. Therefore, our action module also adopts this "think-then-act" strategy. 
We use a ``think-then-act'' approach, similar to chain-of-thought reasoning, enables simulated users to behave more similarly to real users, following \citet{yang2024oasis}.
Inspired by real-world recommendation platforms, we designed a rich set of user actions that more closely reflect authentic interactions. Unlike previous works~\cite{zhang2024generative, wang2025user},
% \cite{zhang2024generative, wang2025user, bougie2025simuser, liu2024lmagent}
our simulated users exhibit interactive actions that can dynamically alter the attributes $A$ of items within the system. The action space includes: do nothing, like product, dislike product, share product, purchase prodcut, create review, like review, and dislike review.

\subsection{Interactive Platform Environment}
\subsubsection{Interaction Implement}
% Previous studies \cite{lee2021product, cui2012effect, li2019effect} have demonstrated that item attributes such as sales, number of likes, and user reviews significantly influence user decision-making. This highlights the importance of modeling the evolution of item attributes based on user feedback in recommendation system simulations. If users cannot interact with the recommendation system during the simulation, time becomes irrelevant—an item recommended at different time points would appear identical to users, which contradicts real-world dynamics. Without incorporating an interaction mechanism, the simulation reduces to a mere aggregation of individual user feedback, lacking the ability to capture the system's macro-level evolution.

To achieve a more realistic simulation of recommender systems, we incorporate an interaction mechanism into \textbf{RecInter}. Specifically, we implement a set of database tables associated with items, which store their dynamic attributes $A$. The agents’ actions are enhanced such that each action $a$ can update these tables in real time. At each recommendation time step, the platform queries the most recent item attributes $A$ from the database and presents them to the simulated users. This method allows the platform to update items dynamically based on user feedback and enable interaction.
% In the experimental section, we compare simulations with and without the interaction mechanism, thereby highlighting the essential role of modeling user-system interactions in accurately capturing the dynamics of recommendation systems.

\subsubsection{Merchant Reply}
In real-world recommendation platforms, merchants also can make changes to the attributes of their items, such as responding to user reviews or modifying product descriptions. To better simulate this interactive environment, we incorporate merchant agent into \textbf{RecInter}. The merchant agent autonomously updates item attributes based on its own strategy and interacts with users through reviews. This addition enables the study of merchant behavior within the recommender systems. 
% For instance, it allows us to investigate how different merchant responses to user reviews affect item sales, and to identify which merchant behaviors are most effective in boosting item performance.

\subsubsection{Recommendation Algorithm}
Recommendation algorithms also constitute a critical component in the simulation of recommender systems. In \textbf{RecInter}, we have integrated a variety of algorithms, including random, most popular, LightGCN~\cite{he2020lightgcn}, MultVAE~\cite{liang2018variational}, and MF~\cite{koren2009matrix}. This module is designed to be flexible and extensible, allowing for the incorporation of custom recommendation algorithms as well. These algorithms aim to recommend items that users are likely to be interested in, thereby enhancing user satisfaction and engagement within the simulation.

\subsection{Behavior Simulation Training}
% Recent works~\cite{gao2023s3, huang2024social, lu2025beyond} have explored fine-tuning models using high-quality simulated data, demonstrating improved capabilities in agent modeling. 
% Inspired by these approaches, 
We adopt a Chain-of-Thought (CoT) fine-tuning approach to enhance the reliability of agent simulation in recommender systems. To construct our training dataset, we used GPT-4o as the base model and ran \textbf{RecInter} multiple times, guiding the model to ``think-then-act''. This process generated a substantial number of simulated interactions enriched with CoT reasoning. To ensure the quality of the data, we implemented a multi-stage filtering pipeline consisting of four key components: (1) Format Filter: Ensures that the model outputs conform to the required structural format. (2) Preference Filter: Verifies alignment between the agent's simulated actions and the user's actual preferences by leveraging real user interaction data. Specifically, we check whether positively interacted items appeared in the user’s real interactions and whether negatively interacted items did not. (3) LLM Filter: Utilizes LLM to assess the plausibility and consistency of the simulated outputs. (4) Human Filter: Involves manual verification to further ensure data quality.

Following this pipeline, we curated a high-quality dataset comprising 5,295 CoT enhanced samples. We then fine-tuned the Qwen-2.5-7B-Instruct model, resulting in the base model that achieved the best simulation performance. The fine-tuning setting is provided in the Appendix~\ref{app:fine-tuning_setting}.
% Experimental results show that our CoT fine-tuning strategy significantly improves the fidelity and reliability of agent behavior simulation.

\section{Experiment}
\subsection{Simulation Credibility}
\textbf{Settings} We employ the fine-tuned Qwen2.5-7B-Instruct as the default base model for \textbf{RecInter}. By default, the recommendation algorithm used is LightGCN, with 10 time steps and 1,000 user agents. The GoogleLocal serves as the default dataset. Additional experimental settings and variations will be specified in subsequent each sections.

\noindent \textbf{Baselies} Our baselines include RecAgent~\cite{wang2025user}, Agent4Rec~\cite{zhang2024generative}, SimUSER~\cite{bougie2025simuser}. Please refer to Appendix~\ref{baseline} for more details.

\subsubsection{Metric-Based Evaluation}
\label{sec:metic_based_evaluation}
\begin{table*}[htbp]
  \centering
  \resizebox{\textwidth}{!}{
    \begin{tabular}{c|cccc|cccc|cccc}
    \toprule
    \multirow{2}[2]{*}{\textbf{Method}} & \multicolumn{4}{c|}{\textbf{GoogleLocal}} & \multicolumn{4}{c|}{\textbf{MovieLens}} & \multicolumn{4}{c}{\textbf{AmazonBook}} \\
          & \textbf{Accuracy} & \textbf{Precision} & \textbf{Recall} & \textbf{F1 Score} & \textbf{Accuracy} & \textbf{Precision} & \textbf{Recall} & \textbf{F1 Score} & \textbf{Accuracy} & \textbf{Precision} & \textbf{Recall} & \textbf{F1 Score} \\
    \midrule
    \textbf{RecAgent (1:1)} & 0.5643  & 0.5832  & 0.5342  & 0.5576  & 0.5807  & 0.6391  & 0.6035  & 0.6205  & 0.6035  & 0.6539  & 0.6636  & 0.6587  \\
    \textbf{RecAgent (1:3)} & 0.5012  & 0.6134  & 0.3765  & 0.4666  & 0.5077  & 0.7396  & 0.3987  & 0.5181  & 0.6144  & 0.6676  & 0.4001  & 0.5003  \\
    \textbf{RecAgent (1:9)} & 0.4625  & 0.6213  & 0.1584  & 0.2523  & 0.4800  & 0.7491  & 0.2168  & 0.3362  & 0.6222  & 0.6641  & 0.1652  & 0.2647  \\
    \midrule
    \textbf{Agent4Rec (1:1)} & 0.6281  & 0.6134  & 0.6223  & 0.6178  & 0.6912  & 0.7460  & 0.6914  & 0.6982  & 0.7190  & 0.7276  & 0.7335  & 0.7002  \\
    \textbf{Agent4Rec (1:3)} & 0.6012  & 0.6456  & 0.3905  & 0.4866  & 0.6675  & 0.7623  & 0.4210  & 0.5433  & 0.6707  & 0.6909  & 0.4423  & 0.5098  \\
    \textbf{Agent4Rec (1:9)} & 0.5786  & 0.6631  & 0.2042  & 0.3112  & 0.6175  & 0.7753  & 0.2139  & 0.3232  & 0.6617  & 0.6939  & 0.2369  & 0.3183  \\
    \midrule
    \textbf{SimUSER (1:1)} & 0.6823  & 0.6312  & 0.6754  & 0.6526  & 0.7912  & 0.7976  & 0.7576  & 0.7771  & 0.8221  & 0.7969  & 0.7841  & 0.7904  \\
    \textbf{SimUSER (1:3)} & 0.6489  & 0.6624  & 0.3893  & 0.4904  & 0.7737  & 0.8173  & 0.5223  & 0.6373  & 0.6629  & 0.7547  & 0.5657  & 0.6467  \\
    \textbf{SimUSER (1:9)} & 0.6042  & 0.6923  & 0.2187  & 0.3324  & 0.6791  & 0.8382  & 0.3534  & 0.4972  & 0.6497  & 0.7588  & 0.3229  & 0.4530  \\
    \midrule
    \textbf{RecInter(1:1)} & \textbf{0.7143 } & 0.6646  & \textbf{0.7057 } & \textbf{0.6854 } & \textbf{0.7947 } & 0.8092  & \textbf{0.7595 } & \textbf{0.7812 } & \textbf{0.8302 } & \textbf{0.8049 } & \textbf{0.7901 } & \textbf{0.7975 } \\
    \textbf{RecInter(1:3)} & 0.6753  & 0.7038  & 0.4312  & 0.5357  & 0.7852  & 0.8236  & 0.5474  & 0.6476  & 0.6804  & 0.7651  & 0.5813  & 0.6614  \\
    \textbf{RecInter(1:9)} & 0.6218  & \textbf{0.7580 } & 0.2508  & 0.3769  & 0.6869  & \textbf{0.8391 } & 0.3638  & 0.5054  & 0.6634  & 0.7631  & 0.3214  & 0.4547  \\
    \bottomrule
    \end{tabular}%
    }
  
  \caption{Metric-based comparison of simulation credibility across different methods with the best results highlighted in bold. Our approach achieves superior performance compared to baselines across all datasets.}
  \vspace{-0.4cm}
  \label{tab:metric_eval}%
\end{table*}%

This part of the experimental setup follows the evaluation used in \citet{zhang2024generative}. In the experiments, 1,000 agents provide feedback on 20 random items. These 20 items consist of both really interacted and non-interacted items by the users, mixed in a $1:m$ ratio. Each agent selects the items they are interested in. Based on the correctness of the agents' selections, we compute evaluation metrics such as Accuracy, Precision, Recall, and F1 Score. 
% Correct choices made by users indicate behavior that closely simulates real user responses. 
Experiments are conducted on three real-world datasets: Google Local~\cite{li2022uctopic}, MovieLens~\cite{harper2015movielens}, and AmazonBook~\cite{mcauley2015image}. The preprocessing steps for datasets are described in the Appendix~\ref{dataset_preprocess}. All experimental results are showed in Table~\ref{tab:metric_eval}. The experimental results demonstrate that our approach outperforms existing methods across all datasets, indicating that the proposed Multidimensional User Profiling and Behavior Simulation Training significantly enhance the accuracy of user simulation in recommender systems. 
% These findings provide a reliable foundation for further experiments and research in the area of recommender systems simulations.

\subsubsection{LLM-Based Evaluation}
To address the limitations of metric-based evaluatios in capturing complex agent behaviors, we introduce an LLM-based evaluation that assesses performance across the entire simulation process. The detailed experimental procedures are provided in the Appendix~\ref{llm_eval}. The complete results are presented in Table~\ref{tab:llm_eval}. As shown, our method achieves a most higher Adjusted Win Rate of 0.6917. 
% This result further validates the superior Simulation Credibility of our method, indicating that the simulated agent behavior more closely aligns with real user logic and corresponding user profiles. 
To further validate the reliability of our LLM-based evaluation method, we conducted a human evaluation study in Appendix~\ref{human_study}. 

\subsubsection{Macro-Level Evaluation}
In addition to assessing the simulation credibility from the perspective of the user agent, we further evaluated the overall credibility of the simulation by comparing the alignment between the real data and simulation results in terms of the distribution of actions and items.

\begin{figure}
\centering
\includegraphics[width=0.9\linewidth]{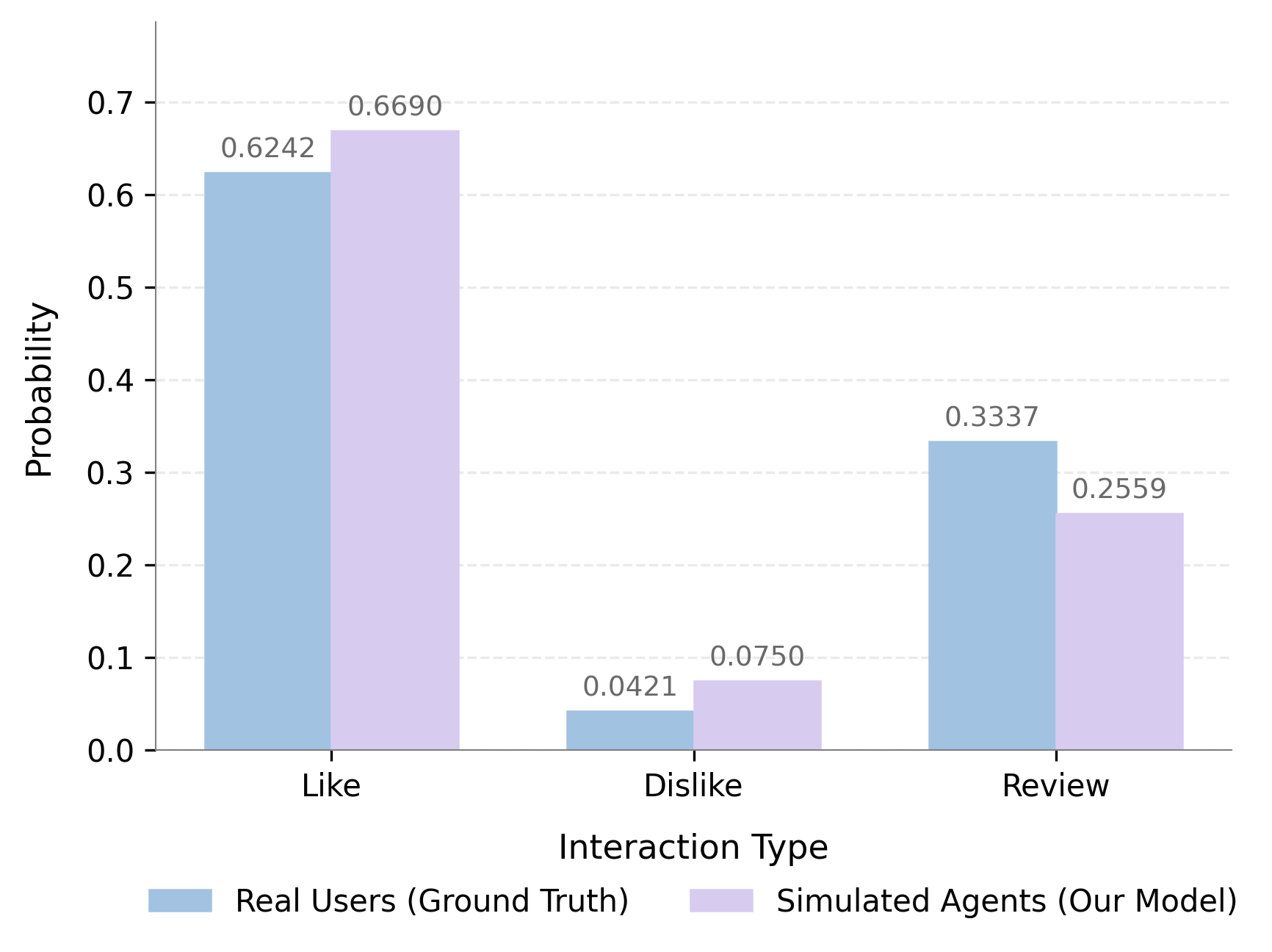}
\caption{Comparison of actions distribution.}
\label{fig:actions_dis}
\vspace{-0.4cm}
\end{figure}

\noindent \textbf{Actions Distribution}
We conducted a comparative analysis of behavioral distributions between real users and simulated agents across three representative actions: like, dislike, and review. Real user data was extracted from the GoogleLocal dataset, while simulated agent data was obtained by executing a full simulation process and then collecting the corresponding actions statistics. As shown in Figure~\ref{fig:actions_dis}, the distribution patterns between the two groups are generally aligned.

\noindent \textbf{Items Distribution}
We also examined differences in items popularity between real users and simulated agents. For real users, we identified the top-10 most popular items from the GoogleLocal dataset and computed their frequency distributions. Similarly, for the simulated agents, we analyzed the top-10 most frequently interacted items based on the simulation results. As shown in Figure~\ref{fig:items_dis}, five out of the top six most popular items overlapped between the two groups, and the overall popularity distributions showed a high degree of similarity.

\subsection{Impact of Interaction Mechanism}
This part of the experiment shows that introducing an interaction mechanism significantly affects the evolutionary process of recommender system simulation, highlighting its indispensable role in the simulation.
In Section~\ref{sec:impact_of_interaction} and Section~\ref{sec:impact_of_malicious_interaction}, we investigate the impact of the presence or absence of interaction and malicious interaction on the simulation evolutionary process. In the Appendix~\ref{merchant_reply}, we demonstrate that the interactive attitude of merchants also shapes this process.

\subsubsection{Impact of the Presence or Absence of Interaction}
\label{sec:impact_of_interaction}
\begin{figure}
\centering
\includegraphics[width=\linewidth]{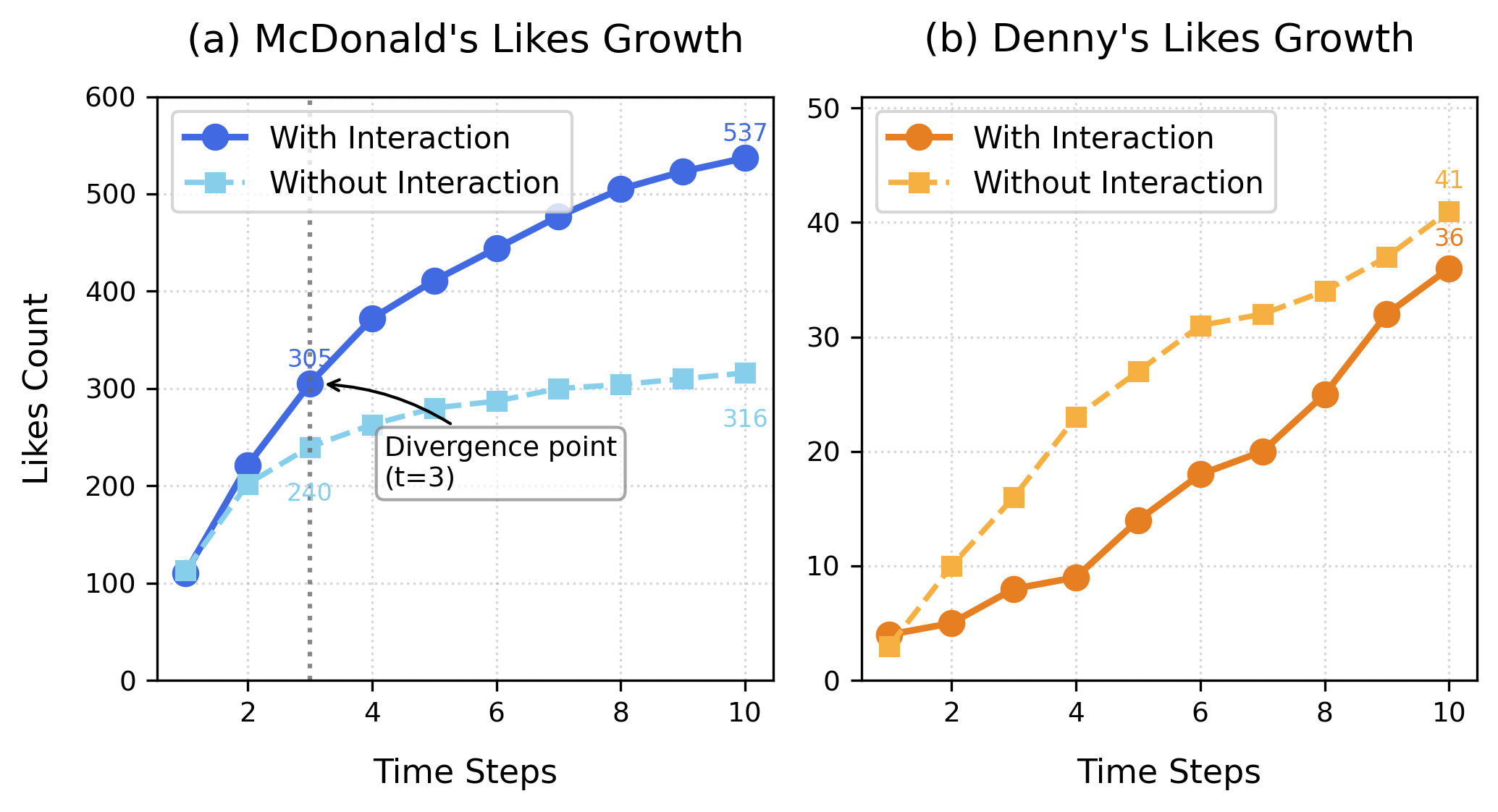}
\caption{Impact of interaction mechanism on likes.}
\label{fig:inter}
\vspace{-0.4cm}
\end{figure}

We conducted two simulations: one incorporating an interaction mechanism and the other without it, while keeping all other settings same. The goal was to observe differences in the simulated evolutionary process between the two simulations, focusing on the changes in the number of likes received by two restaurants—McDonald’s and Denny’s. McDonald’s represents the most popular option, while Denny’s is considered moderately popular. The statistical results are presented in the Figure~\ref{fig:inter}.

The findings indicate a significant difference in the likes for McDonald’s under the two simulation conditions. In the simulation with the interaction mechanism, McDonald’s received a substantially higher number of likes by the end of the simulation. More specifically, from time step t = 3 onward, the increase in likes for McDonald’s became notably greater in the interaction-enabled simulation. This can be attributed to a surge in likes, sales, and reviews for McDonald’s at that point, which likely influenced agents to favor McDonald’s more frequently. In contrast, the difference between the two simulations for Denny’s was relatively minor. This suggests that Denny’s had insufficient attention in the early stages of the simulation, thereby exerting limited influence on the later stages. Additionally, a case study on agent responses in the Appendix~\ref{agent_response_case} further demonstrates that the interaction influences user decision-making during the simulation.

\subsubsection{Impact of Malicious Interaction}
\label{sec:impact_of_malicious_interaction}
\begin{figure*}
\centering
\includegraphics[width=\textwidth]{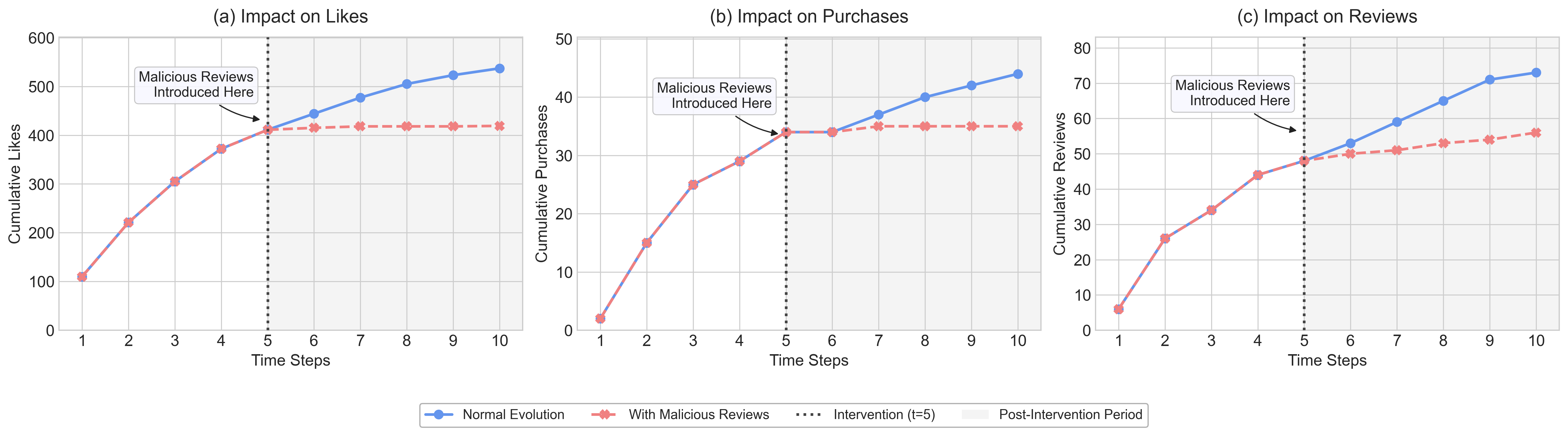}
\caption{Impact of malicious interaction on the cumulative number of likes for McDonald's.}
\vspace{-0.4cm}
\label{fig:mali}
\end{figure*}

We further investigated the impact of malicious interaction on the evolution of the simulation. Specifically, at time step $t=5$, we introduced three malicious reviews targeting McDonald's. By comparing the trends in user likes, purchases, and reviews with or without the introduction of these reviews, we aimed to assess their influence on user behavior. The results are presented in the Figure~\ref{fig:mali}. Compared to the scenario without malicious reviews, the inclusion of such reviews led to a noticeable deceleration in the growth of likes, purchases, and reviews. These findings suggest that malicious reviews can significantly influence user decision-making, which aligns with real-world observations.

\subsection{Ablation Study}

In the ablation study, we evaluate the contributions of the Multidimensional User Profiling and Behavior Simulation Training modules. The experimental setup follows the same configuration as described in Section~\ref{sec:metic_based_evaluation} with the parameter $m$ set to 1. The w/o personalization variant directly uses LLM to summarize the user's profile from 60 sampled history interactions. The w/o training variant employs an untrained Qwen-2.5-7B-Instruct as the agent. As shown in the Table~\ref{tab:ablation}, \textbf{RecInter} achieves the best performance, indicating that both Multidimensional User Profiling and Behavior Simulation Training play critical roles in enhancing the realism of user behavior simulation. In the Appendix~\ref{base_model}, we conduct ablation studies by replacing different base models, demonstrating that our fine-tuned model achieves the best performance. Furthermore, Appendix~\ref{rec_algorithm} evaluate the impact of different recommendation algorithms and show that a effective recommendation strategy significantly enhances the engagement of simulated users on the platform.

\begin{table}[htbp]
  \centering
  \resizebox{\columnwidth}{!}{
    \begin{tabular}{ccccc}
    \toprule
    \textbf{Method } & \textbf{Accuracy} & \textbf{Precision} & \textbf{Recall} & \textbf{F1 Score} \\
    \midrule
    \textbf{w/o personalization} & 0.5733  & 0.5865  & 0.5641  & 0.5601  \\
    \textbf{w/o training} & 0.6715  & 0.6229  & 0.6732  & 0.6471  \\
    \textbf{RecInter} & \textbf{0.7143}  & \textbf{0.6646}  & \textbf{0.7057}  & \textbf{0.6854}  \\
    \bottomrule
    \end{tabular}%
    }
\caption{Ablation study results over 2 variants.}
\vspace{-0.4cm}
  \label{tab:ablation}%
\end{table}%

\subsection{Phenomenon Observation}
In Section~\ref{sec:brand_loyalty} and Section~\ref{sec:matthew_effect}, we reproduced the phenomena of Brand Loyalty and Matthew Effect. We also analyzed Conservative Behavior phenomenon through the reviews of simulated user in the Appendix~\ref{Conservative Behavior}.
\subsubsection{Brand Loyalty}
\label{sec:brand_loyalty}
\begin{figure}
\centering
\includegraphics[width=0.76\linewidth]{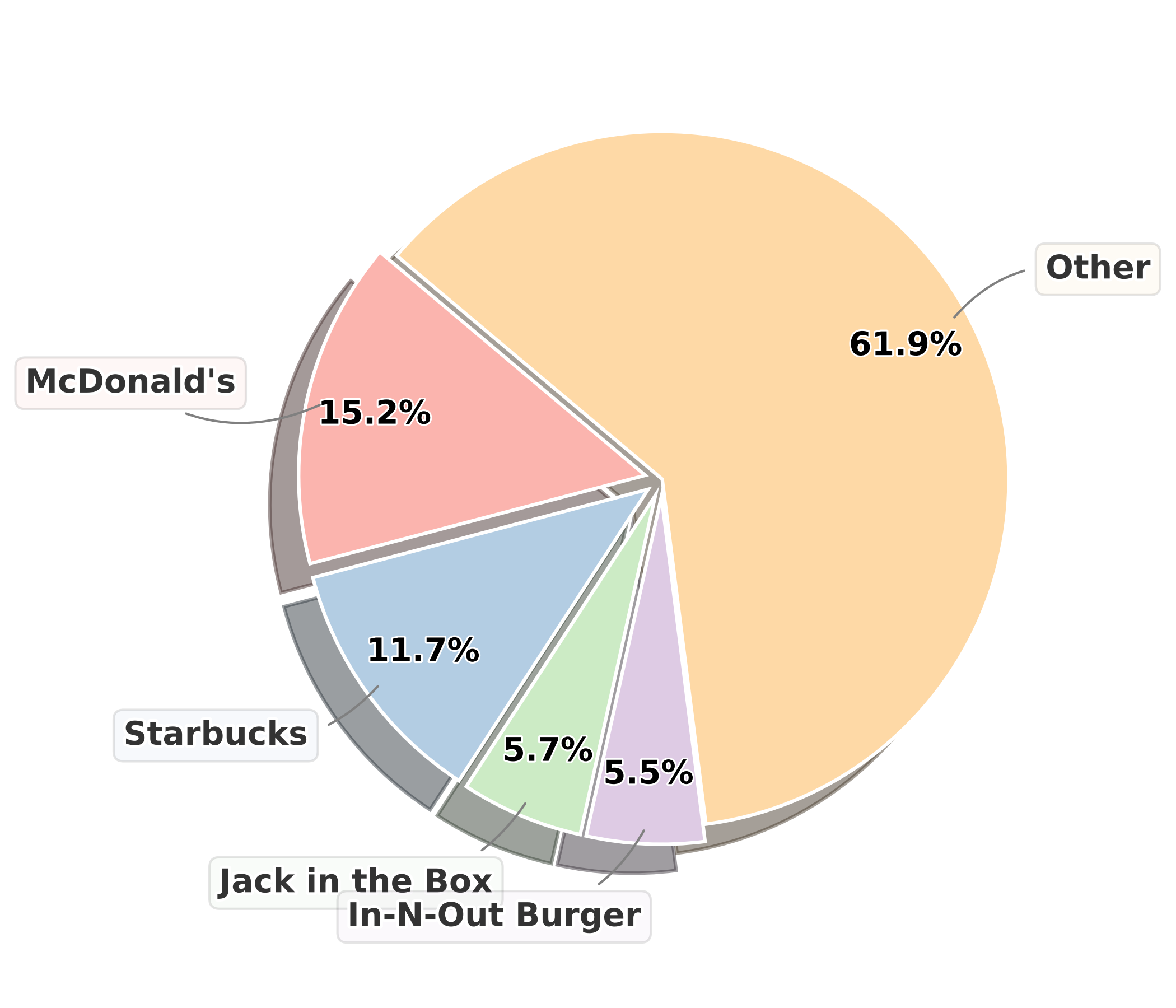}
\caption{Distribution of user interactions across items.}
\vspace{-0.4cm}
\label{fig:brand_pie}
\end{figure}

\begin{figure}
\centering
\includegraphics[width=\linewidth]{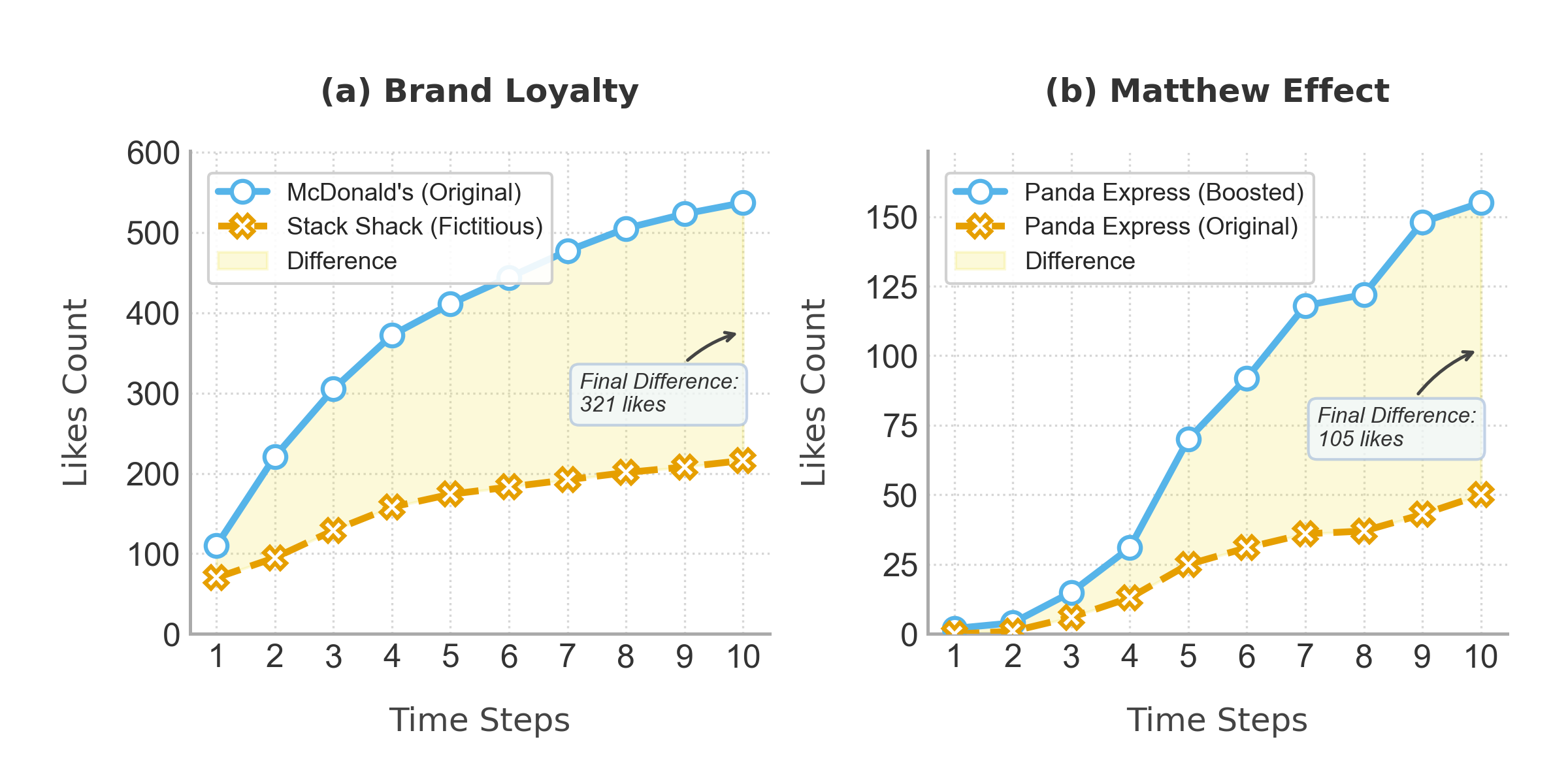}
\caption{The phenomenon observation of Brand Loyalty and Matthew Effect.}
\vspace{-0.4cm}
\label{fig:brand_matthew}
\end{figure}

We analyzed the proportion of interactions each item received at the final stage of the simulation relative to the total number of interactions, as shown in Figure~\ref{fig:brand_pie}. The results indicate that brand-related items were significantly more popular, with McDonald’s and Starbucks accounting for 15.2\% and 11.7\% of all interactions respectively. This suggests the presence of Brand Loyalty among the simulated users. To further investigate this observed Brand Loyalty, we conducted an additional experiment. Specifically, we replaced the brand name ``McDonald’s'' with a fictitious name ``Stack Shack'' while keeping all other attributes of the item, including its recommendation probability, unchanged. The purpose was to examine whether altering the brand name would influence user's choice. As shown in Figure~\ref{fig:brand_matthew}(a), the number of likes for the item decreased substantially after the brand modification.

\subsubsection{Matthew Effect}
\label{sec:matthew_effect}
In Section~\ref{sec:impact_of_interaction}, we showed that adding an interaction mechanism makes users more likely to choose items that received positive feedback earlier. This makes those items more popular over time, which is an example of the Matthew Effect. To further test this effect in our simulation, we did another experiment. We chose Panda Express because it had a medium level of popularity. Before the simulation started, we gave it three positive reviews and set its initial sales to 100. We wanted to see if this would increase its number of likes. As shown in the Figure~\ref{fig:brand_matthew}(b), this manual boost did lead to more likes for Panda Express. This shows that users prefer popular items, supporting the idea that the Matthew Effect exists in our simulation.

\section{Conclusion}
We introduced \textbf{RecInter}, an agent-based simulation platform featuring a novel interaction mechanism. In \textbf{RecInter}, User actions and merchant replies dynamically reshape item attributes, addressing a critical gap in prior simulations. By integrating Multidimensional User Profiling, advanced agent architecture, and Chain-of-Thought fine-tuned LLM, \textbf{RecInter} achieves significantly improved simulation credibility. Our experiments highlight that this dynamic interaction is pivotal for realistically modeling system evolution and observing emergent phenomena like Brand Loyalty and the Matthew Effect. These capabilities position \textbf{RecInter} as a valuable and credible testbed for recommender systems research.

\section{Limitations \& Potential Risks}
Several limitations warrant consideration regarding the current platform's capabilities. Firstly, the depth of user profiling, while multidimensional, relies on LLMs for generating subjective and inferred profiles. Despite efforts to mitigate hallucinations, the inherent biases and comprehension limitations of LLMs may affect the accuracy and completeness of these profiles, potentially failing to fully capture the nuanced complexity of human preferences. Secondly, the user agents operate within a predefined action space (e.g., like, review, purchase). Although this set of actions is rich, it remains fixed, whereas real-world user behavior can be more creative, emergent, or extend beyond these predefined categories. Lastly, the platform's interaction model primarily focuses on user-item, merchant-item, and merchant-user dynamics. Consequently, other significant forms of interaction prevalent in real-world recommendation ecosystems, such as user-to-user social influence, community dynamics, or the impact of external events, are not explicitly modeled in the current implementation. Bias propagation remains a risk. Biases present in these sources (e.g., demographic underrepresentation in historical data, inherent biases in the pre-trained LLMs) could be inadvertently amplified within the simulation.

\section*{Acknowledgments}
This work is supported by Meituan through Agentic system X Program. This work is also supported by the Public Computing Cloud, Renmin University of China and by fund for building worldclass universities (disciplines) of Renmin University of China.

\bibliography{custom}

\begin{thebibliography}{40}
\providecommand{\natexlab}[1]{#1}

\bibitem[{Afsar et~al.(2023)Afsar, Crump, and Far}]{reinforcement_survey_simulator}
Mohammad~Mehdi Afsar, Trafford Crump, and Behrouz~H. Far. 2023.
\newblock Reinforcement learning based recommender systems: {A} survey.
\newblock \emph{{ACM} Comput. Surv.}, 55(7):145:1--145:38.

\bibitem[{Bougie and Watanabe(2025)}]{bougie2025simuser}
Nicolas Bougie and Narimasa Watanabe. 2025.
\newblock Simuser: Simulating user behavior with large language models for recommender system evaluation.
\newblock \emph{arXiv preprint arXiv:2504.12722}.

\bibitem[{Chen et~al.(2019)Chen, Li, Li, Jiang, Qi, and Song}]{adver_user_model_simulator}
Xinshi Chen, Shuang Li, Hui Li, Shaohua Jiang, Yuan Qi, and Le~Song. 2019.
\newblock Generative adversarial user model for reinforcement learning based recommendation system.
\newblock In \emph{{ICML}}. {PMLR}.

\bibitem[{Cui et~al.(2012)Cui, Lui, and Guo}]{cui2012effect}
Geng Cui, Hon-Kwong Lui, and Xiaoning Guo. 2012.
\newblock The effect of online consumer reviews on new product sales.
\newblock \emph{International Journal of Electronic Commerce}, 17(1):39--58.

\bibitem[{Fu et~al.(2024)Fu, Ng, Jiang, and Liu}]{fu2024gptscore}
Jinlan Fu, See~Kiong Ng, Zhengbao Jiang, and Pengfei Liu. 2024.
\newblock Gptscore: Evaluate as you desire.
\newblock In \emph{Proceedings of the 2024 Conference of the North American Chapter of the Association for Computational Linguistics: Human Language Technologies (Volume 1: Long Papers)}, pages 6556--6576.

\bibitem[{Harper and Konstan(2015)}]{harper2015movielens}
F~Maxwell Harper and Joseph~A Konstan. 2015.
\newblock The movielens datasets: History and context.
\newblock \emph{Acm transactions on interactive intelligent systems (tiis)}, 5(4):1--19.

\bibitem[{He et~al.(2020)He, Deng, Wang, Li, Zhang, and Wang}]{he2020lightgcn}
Xiangnan He, Kuan Deng, Xiang Wang, Yan Li, Yongdong Zhang, and Meng Wang. 2020.
\newblock Lightgcn: Simplifying and powering graph convolution network for recommendation.
\newblock In \emph{Proceedings of the 43rd International ACM SIGIR conference on research and development in Information Retrieval}, pages 639--648.

\bibitem[{Huang et~al.(2020)Huang, Oosterhuis, de~Rijke, and van Hoof}]{debias_simulator}
Jin Huang, Harrie Oosterhuis, Maarten de~Rijke, and Herke van Hoof. 2020.
\newblock Keeping dataset biases out of the simulation: {A} debiased simulator for reinforcement learning based recommender systems.
\newblock In \emph{RecSys}.

\bibitem[{Ie et~al.(2019)Ie, Hsu, Mladenov, Jain, Narvekar, Wang, Wu, and Boutilier}]{RecSim}
Eugene Ie, Chih{-}Wei Hsu, Martin Mladenov, Vihan Jain, Sanmit Narvekar, Jing Wang, Rui Wu, and Craig Boutilier. 2019.
\newblock Recsim: {A} configurable simulation platform for recommender systems.
\newblock \emph{CoRR}, abs/1909.04847.

\bibitem[{Koren et~al.(2009)Koren, Bell, and Volinsky}]{koren2009matrix}
Yehuda Koren, Robert Bell, and Chris Volinsky. 2009.
\newblock Matrix factorization techniques for recommender systems.
\newblock \emph{Computer}, 42(8):30--37.

\bibitem[{Lee and Hosanagar(2021)}]{lee2021product}
Dokyun Lee and Kartik Hosanagar. 2021.
\newblock How do product attributes and reviews moderate the impact of recommender systems through purchase stages?
\newblock \emph{Management Science}, 67(1):524--546.

\bibitem[{Li et~al.(2022)Li, Shang, and McAuley}]{li2022uctopic}
Jiacheng Li, Jingbo Shang, and Julian McAuley. 2022.
\newblock Uctopic: Unsupervised contrastive learning for phrase representations and topic mining.
\newblock In \emph{Proceedings of the 60th Annual Meeting of the Association for Computational Linguistics (Volume 1: Long Papers)}, pages 6159--6169.

\bibitem[{Li et~al.(2019)Li, Wu, and Mai}]{li2019effect}
Xiaolin Li, Chaojiang Wu, and Feng Mai. 2019.
\newblock The effect of online reviews on product sales: A joint sentiment-topic analysis.
\newblock \emph{Information \& Management}, 56(2):172--184.

\bibitem[{Liang et~al.(2018)Liang, Krishnan, Hoffman, and Jebara}]{liang2018variational}
Dawen Liang, Rahul~G Krishnan, Matthew~D Hoffman, and Tony Jebara. 2018.
\newblock Variational autoencoders for collaborative filtering.
\newblock In \emph{Proceedings of the 2018 world wide web conference}, pages 689--698.

\bibitem[{Lin et~al.(2023)Lin, Zhao, Zhang, Wu, Ping, and Chen}]{AgentSims}
Jiaju Lin, Haoran Zhao, Aochi Zhang, Yiting Wu, Huqiuyue Ping, and Qin Chen. 2023.
\newblock Agentsims: An open-source sandbox for large language model evaluation.
\newblock \emph{CoRR}, abs/2308.04026.

\bibitem[{Liu et~al.(2023)Liu, Yang, Jia, Zhang, Zhou, Dai, Yang, and Vosoughi}]{Socially_alignment}
Ruibo Liu, Ruixin Yang, Chenyan Jia, Ge~Zhang, Denny Zhou, Andrew~M. Dai, Diyi Yang, and Soroush Vosoughi. 2023.
\newblock Training socially aligned language models in simulated human society.
\newblock \emph{CoRR}, abs/2305.16960.

\bibitem[{Liu et~al.(2024{\natexlab{a}})Liu, Zhou, Guo, Shareghi, Vuli{\'c}, Korhonen, and Collier}]{liu2024aligning}
Yinhong Liu, Han Zhou, Zhijiang Guo, Ehsan Shareghi, Ivan Vuli{\'c}, Anna Korhonen, and Nigel Collier. 2024{\natexlab{a}}.
\newblock Aligning with human judgement: The role of pairwise preference in large language model evaluators.
\newblock \emph{arXiv preprint arXiv:2403.16950}.

\bibitem[{Liu et~al.(2024{\natexlab{b}})Liu, Chen, Zhang, Gao, Zhang, and Yan}]{liu2024skepticism}
Yuhan Liu, Xiuying Chen, Xiaoqing Zhang, Xing Gao, Ji~Zhang, and Rui Yan. 2024{\natexlab{b}}.
\newblock From skepticism to acceptance: Simulating the attitude dynamics toward fake news.
\newblock \emph{arXiv preprint arXiv:2403.09498}.

\bibitem[{Liu et~al.(2025)Liu, Liu, Zhang, Chen, and Yan}]{liu2025truth}
Yuhan Liu, Yuxuan Liu, Xiaoqing Zhang, Xiuying Chen, and Rui Yan. 2025.
\newblock The truth becomes clearer through debate! multi-agent systems with large language models unmask fake news.
\newblock \emph{arXiv preprint arXiv:2505.08532}.

\bibitem[{Liu et~al.(2024{\natexlab{c}})Liu, Song, Zhang, Chen, and Yan}]{liu2024tiny}
Yuhan Liu, Zirui Song, Xiaoqing Zhang, Xiuying Chen, and Rui Yan. 2024{\natexlab{c}}.
\newblock From a tiny slip to a giant leap: An llm-based simulation for fake news evolution.
\newblock \emph{arXiv preprint arXiv:2410.19064}.

\bibitem[{Luo et~al.(2022)Luo, Liu, Xiao, Xie, and Li}]{mindsim_simulator}
Xufang Luo, Zheng Liu, Shitao Xiao, Xing Xie, and Dongsheng Li. 2022.
\newblock Mindsim: User simulator for news recommenders.
\newblock In \emph{{WWW}}.

\bibitem[{McAuley et~al.(2015)McAuley, Targett, Shi, and Van Den~Hengel}]{mcauley2015image}
Julian McAuley, Christopher Targett, Qinfeng Shi, and Anton Van Den~Hengel. 2015.
\newblock Image-based recommendations on styles and substitutes.
\newblock In \emph{Proceedings of the 38th international ACM SIGIR conference on research and development in information retrieval}, pages 43--52.

\bibitem[{Park et~al.(2023{\natexlab{a}})Park, O'Brien, Cai, Morris, Liang, and Bernstein}]{park2023generative}
Joon~Sung Park, Joseph O'Brien, Carrie~Jun Cai, Meredith~Ringel Morris, Percy Liang, and Michael~S Bernstein. 2023{\natexlab{a}}.
\newblock Generative agents: Interactive simulacra of human behavior.
\newblock In \emph{Proceedings of the 36th annual acm symposium on user interface software and technology}, pages 1--22.

\bibitem[{Park et~al.(2023{\natexlab{b}})Park, O'Brien, Cai, Morris, Liang, and Bernstein}]{Generative_Agent}
Joon~Sung Park, Joseph~C. O'Brien, Carrie~J. Cai, Meredith~Ringel Morris, Percy Liang, and Michael~S. Bernstein. 2023{\natexlab{b}}.
\newblock Generative agents: Interactive simulacra of human behavior.
\newblock \emph{CoRR}, abs/2304.03442.

\bibitem[{Qian et~al.(2023)Qian, Cong, Yang, Chen, Su, Xu, Liu, and Sun}]{ChatDev}
Chen Qian, Xin Cong, Cheng Yang, Weize Chen, Yusheng Su, Juyuan Xu, Zhiyuan Liu, and Maosong Sun. 2023.
\newblock Communicative agents for software development.
\newblock \emph{CoRR}, abs/2307.07924.

\bibitem[{Ricci et~al.(2010)Ricci, Rokach, and Shapira}]{ricci2010introduction}
Francesco Ricci, Lior Rokach, and Bracha Shapira. 2010.
\newblock Introduction to recommender systems handbook.
\newblock In \emph{Recommender systems handbook}, pages 1--35. Springer.

\bibitem[{Sachdeva and McAuley(2020)}]{sachdeva2020useful}
Noveen Sachdeva and Julian McAuley. 2020.
\newblock How useful are reviews for recommendation? a critical review and potential improvements.
\newblock In \emph{proceedings of the 43rd international ACM SIGIR conference on research and development in information retrieval}, pages 1845--1848.

\bibitem[{Salton and Buckley(1988)}]{salton1988term}
Gerard Salton and Christopher Buckley. 1988.
\newblock Term-weighting approaches in automatic text retrieval.
\newblock \emph{Information processing \& management}, 24(5):513--523.

\bibitem[{Shi et~al.(2019)Shi, Yu, Da, Chen, and Zeng}]{Virtual-Taobao}
Jing{-}Cheng Shi, Yang Yu, Qing Da, Shi{-}Yong Chen, and Anxiang Zeng. 2019.
\newblock Virtual-taobao: Virtualizing real-world online retail environment for reinforcement learning.
\newblock In \emph{{AAAI}}.

\bibitem[{Srifi et~al.(2020)Srifi, Oussous, Ait~Lahcen, and Mouline}]{srifi2020recommender}
Mehdi Srifi, Ahmed Oussous, Ayoub Ait~Lahcen, and Salma Mouline. 2020.
\newblock Recommender systems based on collaborative filtering using review texts—a survey.
\newblock \emph{Information}, 11(6):317.

\bibitem[{Wang et~al.(2023{\natexlab{a}})Wang, Xie, Jiang, Mandlekar, Xiao, Zhu, Fan, and Anandkumar}]{Voyager}
Guanzhi Wang, Yuqi Xie, Yunfan Jiang, Ajay Mandlekar, Chaowei Xiao, Yuke Zhu, Linxi Fan, and Anima Anandkumar. 2023{\natexlab{a}}.
\newblock Voyager: An open-ended embodied agent with large language models.
\newblock \emph{CoRR}, abs/2305.16291.

\bibitem[{Wang et~al.(2023{\natexlab{b}})Wang, Ma, Feng, Zhang, Yang, Zhang, Chen, Tang, Chen, Lin, Zhao, Wei, and Wen}]{Wang2023survey}
Lei Wang, Chen Ma, Xueyang Feng, Zeyu Zhang, Hao Yang, Jingsen Zhang, Zhiyuan Chen, Jiakai Tang, Xu~Chen, Yankai Lin, Wayne~Xin Zhao, Zhewei Wei, and Ji{-}Rong Wen. 2023{\natexlab{b}}.
\newblock A survey on large language model based autonomous agents.
\newblock \emph{CoRR}, abs/2308.11432.

\bibitem[{Wang et~al.(2025)Wang, Zhang, Yang, Chen, Tang, Zhang, Chen, Lin, Sun, Song et~al.}]{wang2025user}
Lei Wang, Jingsen Zhang, Hao Yang, Zhi-Yuan Chen, Jiakai Tang, Zeyu Zhang, Xu~Chen, Yankai Lin, Hao Sun, Ruihua Song, and 1 others. 2025.
\newblock User behavior simulation with large language model-based agents.
\newblock \emph{ACM Transactions on Information Systems}, 43(2):1--37.

\bibitem[{Wu et~al.(2023)Wu, Bansal, Zhang, Wu, Zhang, Zhu, Li, Jiang, Zhang, and Wang}]{AutoGen}
Qingyun Wu, Gagan Bansal, Jieyu Zhang, Yiran Wu, Shaokun Zhang, Erkang Zhu, Beibin Li, Li~Jiang, Xiaoyun Zhang, and Chi Wang. 2023.
\newblock Autogen: Enabling next-gen {LLM} applications via multi-agent conversation framework.
\newblock \emph{CoRR}, abs/2308.08155.

\bibitem[{Xi et~al.(2025)Xi, Chen, Guo, He, Ding, Hong, Zhang, Wang, Jin, Zhou et~al.}]{xi2025rise}
Zhiheng Xi, Wenxiang Chen, Xin Guo, Wei He, Yiwen Ding, Boyang Hong, Ming Zhang, Junzhe Wang, Senjie Jin, Enyu Zhou, and 1 others. 2025.
\newblock The rise and potential of large language model based agents: A survey.
\newblock \emph{Science China Information Sciences}, 68(2):121101.

\bibitem[{Xu et~al.(2023)Xu, Wang, Li, Luo, Wang, Liu, and Liu}]{WereWolf}
Yuzhuang Xu, Shuo Wang, Peng Li, Fuwen Luo, Xiaolong Wang, Weidong Liu, and Yang Liu. 2023.
\newblock Exploring large language models for communication games: An empirical study on werewolf.
\newblock \emph{CoRR}, abs/2309.04658.

\bibitem[{Yang et~al.(2021)Yang, Dai, Dong, Chen, He, and Wang}]{top_n_simulator}
Mengyue Yang, Quanyu Dai, Zhenhua Dong, Xu~Chen, Xiuqiang He, and Jun Wang. 2021.
\newblock Top-n recommendation with counterfactual user preference simulation.
\newblock In \emph{{CIKM}}.

\bibitem[{Yang et~al.(2024)Yang, Zhang, Zheng, Jiang, Gan, Wang, Ling, Chen, Ma, Dong et~al.}]{yang2024oasis}
Ziyi Yang, Zaibin Zhang, Zirui Zheng, Yuxian Jiang, Ziyue Gan, Zhiyu Wang, Zijian Ling, Jinsong Chen, Martz Ma, Bowen Dong, and 1 others. 2024.
\newblock Oasis: Open agents social interaction simulations on one million agents.
\newblock \emph{arXiv preprint arXiv:2411.11581}.

\bibitem[{Zhang et~al.(2024)Zhang, Chen, Sheng, Wang, and Chua}]{zhang2024generative}
An~Zhang, Yuxin Chen, Leheng Sheng, Xiang Wang, and Tat-Seng Chua. 2024.
\newblock On generative agents in recommendation.
\newblock In \emph{Proceedings of the 47th international ACM SIGIR conference on research and development in Information Retrieval}, pages 1807--1817.

\bibitem[{Zheng et~al.(2023)Zheng, Chiang, Sheng, Zhuang, Wu, Zhuang, Lin, Li, Li, Xing et~al.}]{zheng2023judging}
Lianmin Zheng, Wei-Lin Chiang, Ying Sheng, Siyuan Zhuang, Zhanghao Wu, Yonghao Zhuang, Zi~Lin, Zhuohan Li, Dacheng Li, Eric Xing, and 1 others. 2023.
\newblock Judging llm-as-a-judge with mt-bench and chatbot arena.
\newblock \emph{Advances in Neural Information Processing Systems}, 36:46595--46623.

\end{thebibliography}

\clearpage

\appendix

\section{Experiment Setting}
\subsection{Dataset Preprocessing}
\label{dataset_preprocess}
For the MovieLens~\cite{harper2015movielens} and AmazonBook~\cite{mcauley2015image} datasets, we followed the preprocessing procedures used by Agent4Rec~\cite{zhang2024generative}. In the case of the GoogleLocal~\cite{li2022uctopic} dataset, we first filtered items related to the restaurant domain within the California region. We then selected users with an interaction history longer than 300 and items having more than 10 interactions. The resulting subset constitutes the GoogleLocal dataset used in our experiments.

\subsection{Baseline Details}
\label{baseline}
\noindent \textbf{RecAgent}~\cite{wang2025user}: RecAgent uses LLM-powered agents, each with a profile, memory, and action module, to simulate diverse user behaviors in a recommendation sandbox environment.

\noindent \textbf{Agent4Rec}~\cite{zhang2024generative}: Agent4Rec simulates users for recommendation systems using LLM-powered agents that have profiles derived from real-world data and perform taste and emotion-driven actions.

\noindent \textbf{SimUSER}~\cite{bougie2025simuser}: SimUSER creates user personas from historical data, then having LLM-powered agents equipped with these personas, perception, memory, and a decision-making brain module interact with the system.

\subsection{Specific Hyperparameter Setting in Memory Module}
\label{app:hyperparameter}
In our implementation of the memory module, we set the hyperparameters $\alpha$ and $\beta$ to 0.7 and 0.3, respectively. The value of $\gamma$ is set to 0.2, while $\theta_p$ and $\theta_c$ are set to 25 and 5, respectively.

\subsection{Fine-tuning Setting}
\label{app:fine-tuning_setting}
The fine-tuning phase was conducted using 4 NVIDIA A100 GPUs. The batch size is set to 8 per device and the learning rate in the cosine learning rate scheduler is 1.0e-4 over three epochs. LoRA was applied to all linear modules while LoRA rank is set to 8 and LoRA alpha is set to 16. The dataset was divided into a training set and a validation set at a ratio of 9:1. We select the model that achieves the best performance on the validation set as the final trained model.

\subsection{Base Model Setting}
\label{app:base_model_setting}
For closed-source large language models, we access them via the official API, configuring the temperature to 0, top-p to 1, and setting both the frequency penalty and presence penalty to 0. For open-source models, we deploy them locally using vllm, and the API parameter configurations are consistent with closed-source models.

\section{Supplementary Experiments}
\subsection{Impact of Merchant Reply}
\label{merchant_reply}

Thanks to the design of our interaction mechanism, it becomes feasible to investigate the impact of merchant decision-making on recommendation system simulation. In this experiment, we selected McDonald’s as the focal case due to its high level of popularity. We designed three types of merchant response attitudes: (1) merchant who do not respond to user reviews; (2) merchant who actively engage with users in order to defend their brand and interests; and (3) merchant who respond negatively, potentially engaging in verbal conflicts with users. Our objective is to examine how these three distinct response strategies adopted by McDonald’s influence the likes of its products within the recommendation system simulation.

\begin{figure}
\centering
\includegraphics[width=\linewidth]{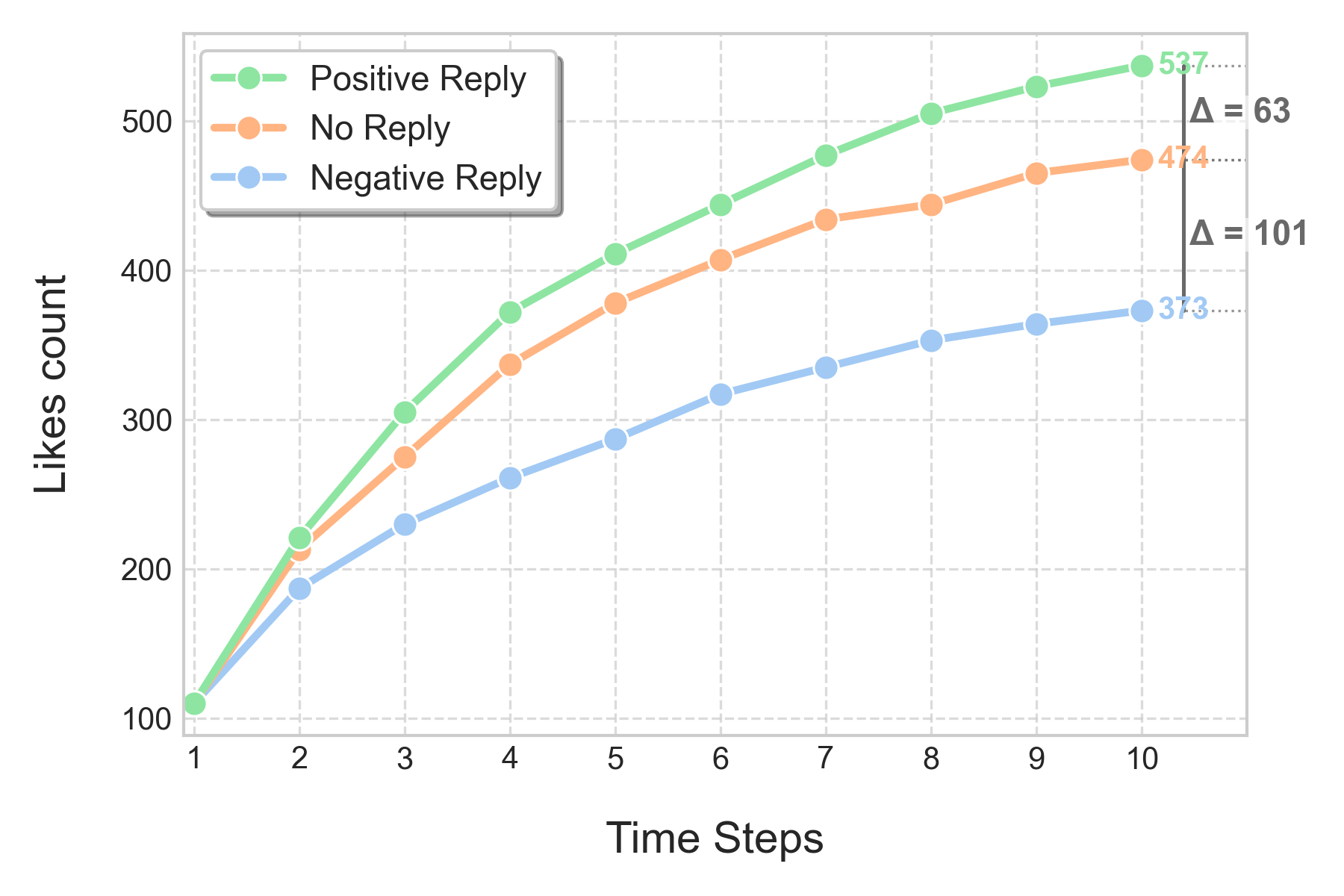}
\caption{Impact of merchant reply on likes count.}
\label{fig:merchant}
\end{figure}

As shown in the Figure~\ref{fig:merchant}, merchants who actively respond to user reviews can significantly increase the number of likes their store receive. At the end of the simulation, the store from merchant with positive responses received 164 and 63 more likes compared to those with negative or no responses, respectively. This significant difference demonstrates that the responsiveness of merchants has a notable impact on the popularity of their store.

\subsection{Conservative Behavior}
\label{Conservative Behavior}
\begin{figure}
\centering
\includegraphics[width=\linewidth]{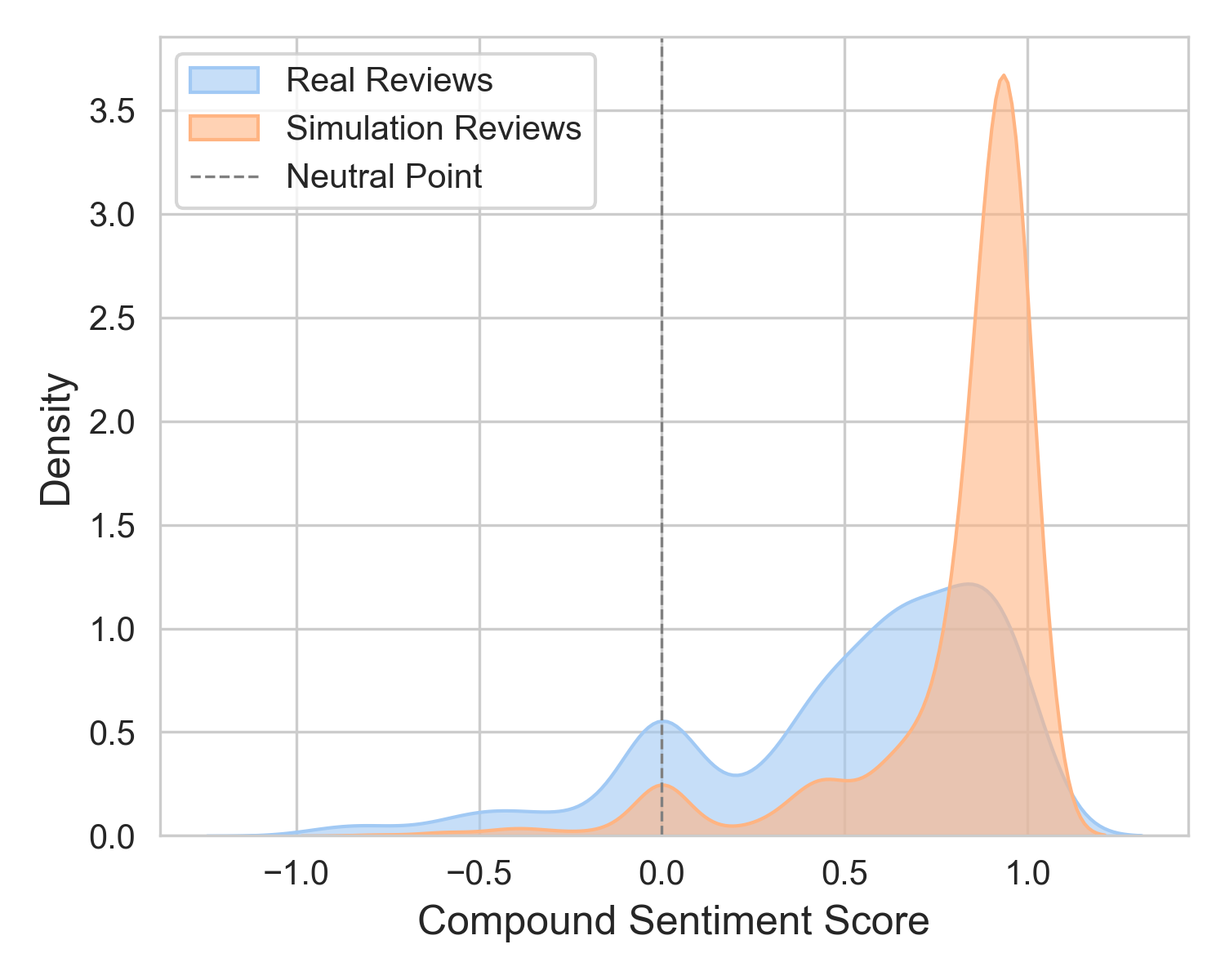}
\caption{Comparison of compound sentiment score between real reviews and simulation reviews.}
\label{fig:review_sentiment}
\end{figure}

\begin{figure}
\centering
\includegraphics[width=\linewidth]{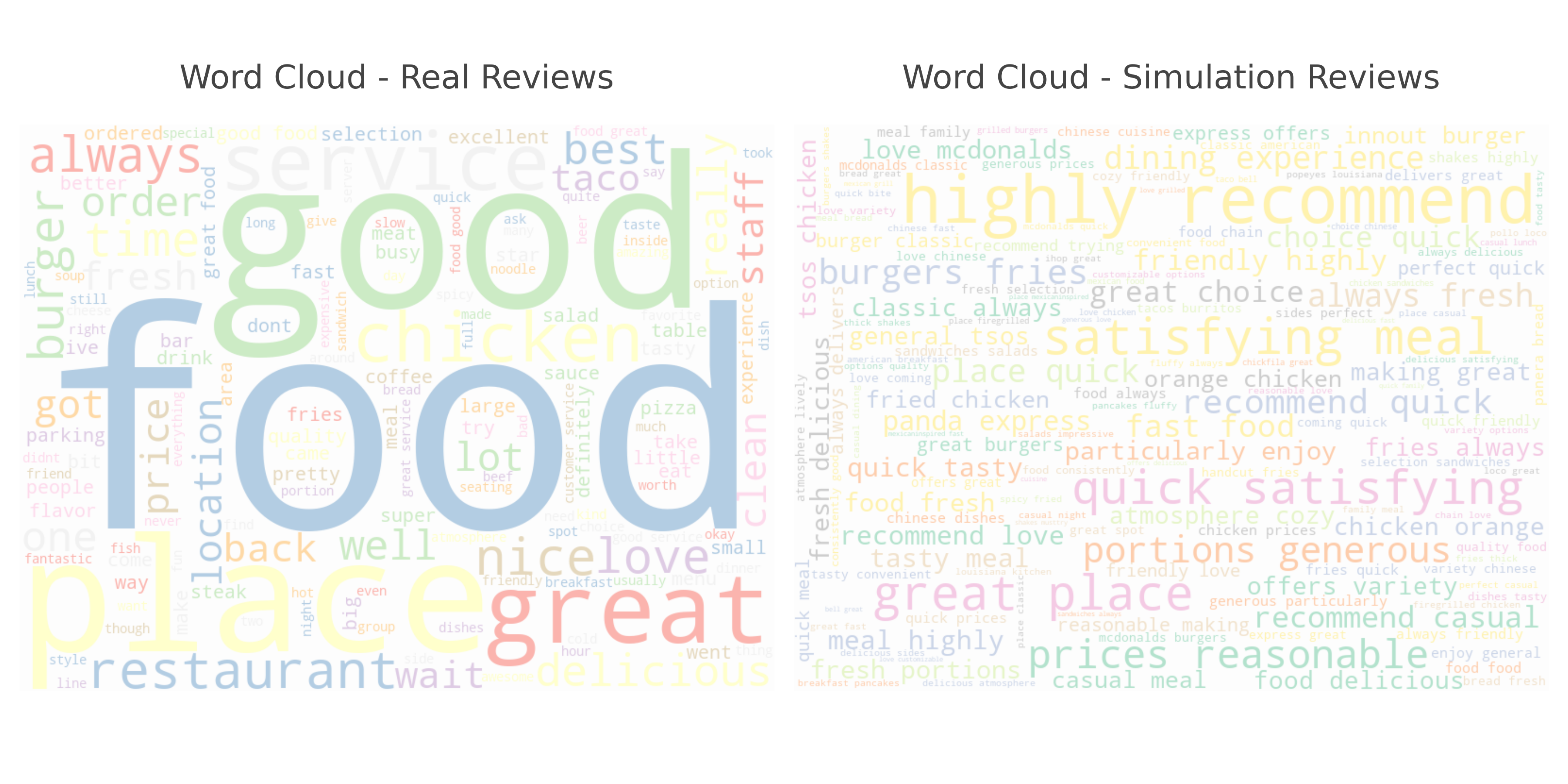}
\caption{The word clouds of real reviews and simulation reviews.}
\label{fig:review_wordcloud}
\end{figure}

We conducted an analysis comparing reviews generated by simulated users and those from real users. By randomly sampling 1,000 reviews from both the real-world dataset and the simulated dataset, we applied the VADER sentiment analysis tool to compute the Compound Sentiment Scores. The results, as shown in the Figure~\ref{fig:review_sentiment}, indicate that both groups of reviews exhibit a similar overall sentiment pattern: predominantly positive, with relatively few negative reviews, and a higher proportion of positive over neutral reviews. However, simulated users demonstrated a significantly higher tendency to leave positive reviews. We attribute this to the ethical alignment mechanisms in large language models, which encourage more cautious and friendly responses, reflecting the conservative behavior of simulated user. Additionally, we generated word clouds for both reviews shown in Figure~\ref{fig:review_wordcloud}. The results reveal that real user reviews tend to be more colloquial and feature concentrated vocabulary, whereas simulated users prefer more formal and structured expressions. Future work could focus on refining simulated user behavior to better mimic authentic user reviews.

\subsection{Items Distribution Comparison}
\label{item_distribution}
\begin{figure}
\centering
\includegraphics[width=0.78\linewidth]{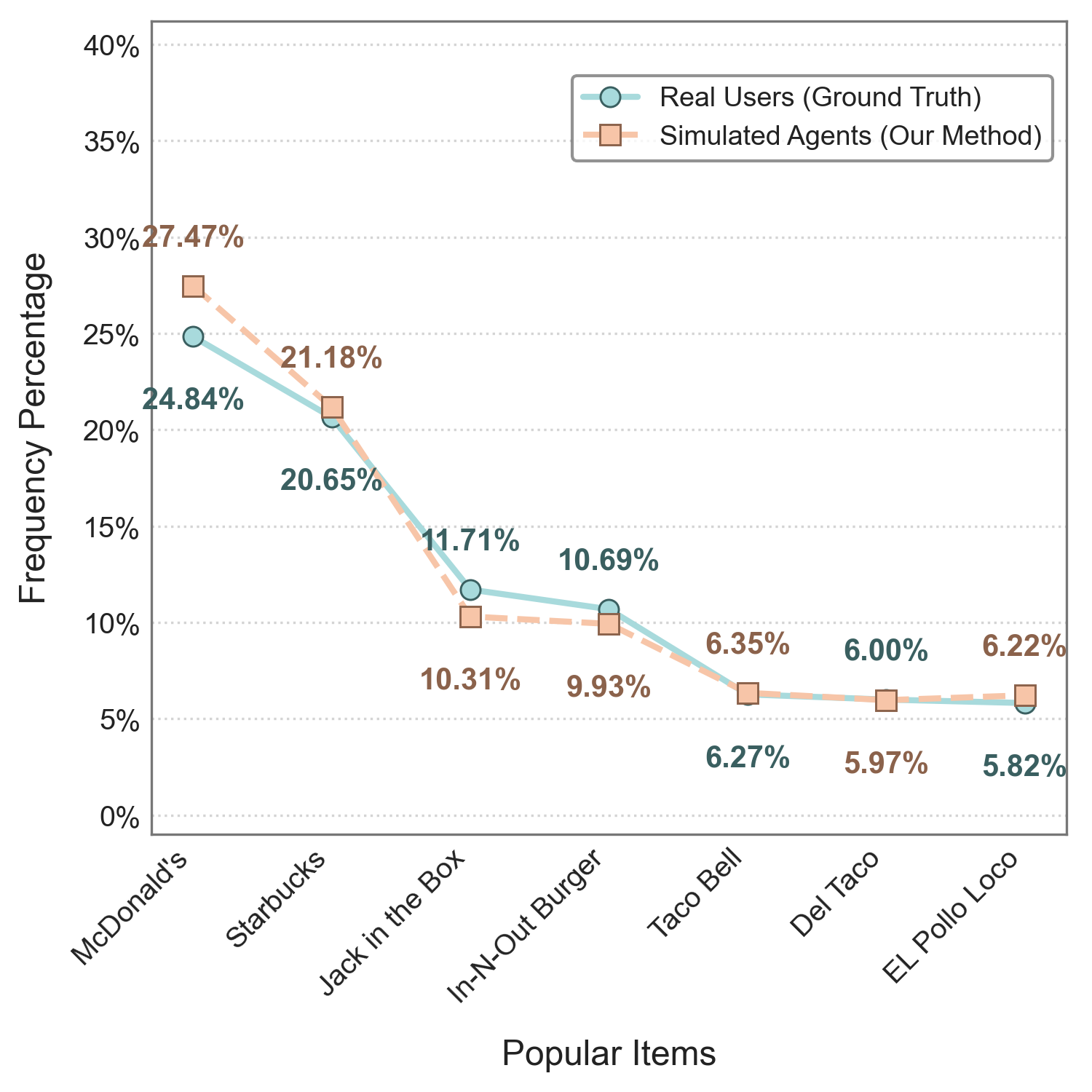}
\caption{Items distribution comparison between reality and simulation for GoogleLocal dataset.}
\label{fig:items_dis}
\vspace{-0.4cm}
\end{figure}

The detailed items distribution comparison between reality and simulation for GoogleLocal dataset is shown in Figure~\ref{fig:items_dis}.

\subsection{Impact of Base Model}
\label{base_model}

We maintained the same configuration as \ref{sec:metic_based_evaluation}, setting $m=1$ and experimented with different base models. The results are presented in Table~\ref{tab:impart_of_base_model}. It can be observed that the choice of base model has a certain impact on the final performance; closed-source models such as GPT-4o generally achieve higher accuracy compared to open-source models. However, our fine-tuned model achieved the best results, indicating that our fine-tuning approach can significantly enhance the simulation accuracy of open-source models.

\begin{table}[htbp]
  \centering
    \resizebox{\columnwidth}{!}{
    \begin{tabular}{ccccc}
    \toprule
    \textbf{Model} & \textbf{Accuracy} & \textbf{Precision} & \textbf{Recall} & \textbf{F1 Score} \\
    \midrule
    \textbf{Llama3-8B-Instruct} & 0.6605  & 0.6138  & 0.6697  & 0.6406  \\
    \textbf{Llama3.1-8B-Instruct} & 0.6696  & 0.6214  & 0.6721  & 0.6458  \\
    \textbf{Qwen2.5-7B-Instruct} & 0.6715  & 0.6229  & 0.6732  & 0.6471  \\
    \textbf{Qwen3-8B} & 0.6824  & 0.6214  & 0.6833  & 0.6509  \\
    \textbf{Gemma2-9B-It} & 0.6632  & 0.6172  & 0.6702  & 0.6426  \\
    \textbf{Qwen2.5-14B-Instruct} & 0.6831  & 0.6342  & 0.6673  & 0.6503  \\
    \textbf{GPT-4o-mini} & 0.6852  & 0.6307  & 0.6833  & 0.6559  \\
    \textbf{GPT-4o} & 0.6924  & 0.6319  & 0.6889  & 0.6592  \\
    \textbf{GPT-4.1} & 0.6921  & 0.6284  & 0.6899  & 0.6577  \\
    \midrule
    \textbf{Llama-3-8B-Instruct(finetuned)} & 0.7128  & 0.6634  & 0.7025  & 0.6824  \\
    \textbf{Qwen2.5-7B-Instruct(finetuned)} & \textbf{0.7143}  & \textbf{0.6646}  & \textbf{0.7057}  & \textbf{0.6845}  \\
    \bottomrule
    \end{tabular}%
    }
\caption{Simulation result of \textbf{RecInter} with different base models.}
  \label{tab:impart_of_base_model}%
\end{table}%

\subsection{Imapct of Recommendation Algorithm}
\label{rec_algorithm}

We investigated the impact of different recommendation algorithms within the \textbf{RecInter} in this experiment. Specifically, we experimented with five recommendation algorithms: Random, Most Popular, MF~\cite{koren2009matrix}, LightGCN~\cite{he2020lightgcn}, and MultVAE~\cite{liang2018variational}. After the simulation ended, we collected statistics on user actions, including the number of Like, Purchase, Review, and Dislike. A higher number of Likes, Purchases, and Reviews, along with fewer Dislikes, is indicative of greater user satisfaction with the platform. The experimental results are presented in the Table~\ref{tab:impart_of_rec}. As observed, LightGCN achieved the best performance in terms of user satisfaction, while the Random algorithm resulted in the poorest performance. Overall, the effectiveness of the recommendation algorithms exhibited a positive correlation with user satisfaction, aligning well with real-world expectations.

\begin{table}[htbp]
  \centering
  \resizebox{\columnwidth}{!}{
    \begin{tabular}{ccccc}
    \toprule
    \textbf{Algorithm} & \textbf{Like } & \textbf{Purchase} & \textbf{Review} & \textbf{Dislike} \\
    \midrule
    \textbf{Random} & 2297  & 127   & 843   & 455 \\
    \textbf{Pop} & 2937  & 195   & 908   & 425 \\
    \textbf{MF} & 3041  & 273   & 1038  & 415 \\
    \textbf{MultVAE} & 3134  & 264   & 1018  & 373 \\
    \textbf{LightGCN} & \textbf{3246} & \textbf{282} & \textbf{1242} & \textbf{364} \\
    \bottomrule
    \end{tabular}%
    }
    \caption{Overall performance of \textbf{RecInter} with different recommendation algorithms.}
  \label{tab:impart_of_rec}%
\end{table}%

\subsection{LLM-Based Evaluation Details}
\label{llm_eval}
We first run a full simulation, randomly sample 100 agents, and retrieve their associated memories to construct representative agent simulation samples. An LLM then evaluates these samples based on behavioral logic and alignment with the user's profile to assess simulation reliability. Following prior work~\cite{zheng2023judging, liu2024aligning, fu2024gptscore}, we adopt a pairwise evaluation strategy using a Judge Agent (GPT-4o). Each pair of samples from different method is evaluated twice with swapped order to reduce bias. A win is counted only if one method is consistently preferred in both orders; otherwise, the result is a tie. Finally, we report the number of Wins, Ties, Losses, and compute the Adjusted Win Rate for comparison:

{\small
\begin{align*}
        \text{Adjusted Win Rate}& = \\
        & \frac{\text{Win Counts} + 0.5 \cdot \text{Tie Counts}}{\text{Win Counts} + \text{Loss Counts} + \text{Tie Counts}}
\end{align*}
}

\begin{table}[htbp]
  \centering
  \resizebox{\columnwidth}{!}{
    \begin{tabular}{ccccccc}
    \toprule
    \textbf{Method} & \textbf{Win} & \textbf{Loss} & \textbf{Tie} & \textbf{Win Rate} & \textbf{Loss Rate} & \textbf{Adjusted Win Rate} \\
    \midrule
    \textbf{RecAgent} & 12    & 216   & 72    & 0.0400  & 0.7200  & 0.1600  \\
    \textbf{Agent4Rec} & 114   & 77    & 109   & 0.3800  & 0.2567  & 0.5617  \\
    \textbf{SimUSER} & 106   & 54    & 140   & 0.3533  & 0.1800  & 0.5867  \\
    \textbf{RecInter} & 143   & 28    & 129   & 0.4767  & 0.0933  & \textbf{0.6917 } \\
    \bottomrule
    \end{tabular}%
    }
    \caption{LLM-Based simulation credibility comparison of different methods.}
  \label{tab:llm_eval}%
\end{table}%

\subsection{Human Evaluation Study for LLM Evaluation}
\label{human_study}
To further substantiate the reliability of our LLM-based evaluation, we conducted a human evaluation study. From the pairs evaluated by the GPT-4o Judge Agent, we randomly selected a subset of 30 distinct pairs of agent simulation samples. These pairs were presented to three human evaluators with expertise in agent-based simulation. The three evaluators were all graduate students conducting research in the field of artificial intelligence. The evaluators were provided with the same instructions and criteria as the Judge Agent: to assess which sample in the pair demonstrated more realistic behavioral logic and better alignment with the associated user profile. Evaluators made independent judgments, and we used a majority vote to determine the human-preferred sample for each pair. We then compared the human consensus judgments with the GPT-4o Judge Agent's decisions for these 30 pairs. As shown in Table~\ref{tab:human_study},We observed a high degree of agreement, with the LLM's judgments aligning with the human majority vote in 26 out of 30 cases (86.7\% agreement). This strong correlation between human and LLM evaluation lends significant credibility to our LLM-based evaluation.

\section{Case Study}

\subsection{Subjective Profile Case}
\label{subjective_profile_case}
\begin{figure}
\centering
\includegraphics[width=\linewidth]{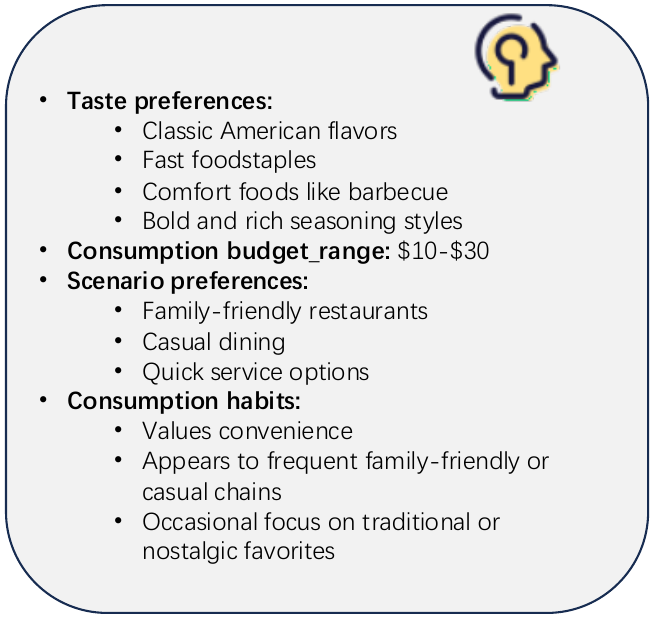}
\caption{A case of Subjective Profile.}
\label{fig:subjective_profile_case}
\end{figure}

A case of Subjective Profile is shown in the Figure~\ref{fig:subjective_profile_case}.

\subsection{Inferred Profile Case}
\label{inferred_profile_case}
\begin{figure}
\centering
\includegraphics[width=\linewidth]{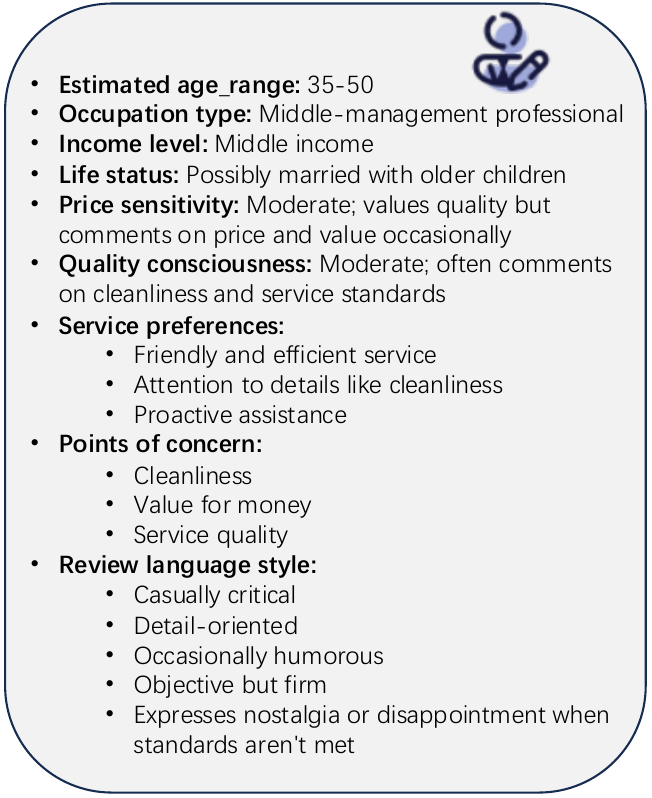}
\caption{A case of Inferred Profile.}
\label{fig:inferred_profile_case}
\end{figure}

A case of Inferred Profile is shown in the Figure~\ref{fig:inferred_profile_case}.

\subsection{Agent Response Case}
\label{agent_response_case}
\begin{figure}
\centering
\includegraphics[width=\linewidth]{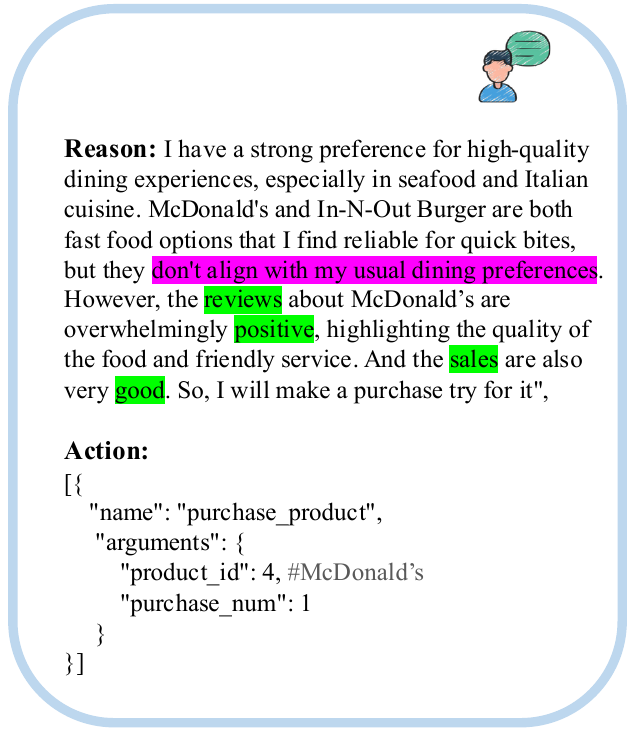}
\caption{A case of agent response.}
\label{fig:agent_response_case}
\end{figure}

We conducted an analysis of the agent responses after completing the full simulation. The Figure~\ref{fig:agent_response_case} illustrates the response case of simulated $426$-th agent at time step 3. Although McDonald’s does not align with the agent’s dining preferences, the agent ultimately decided to make a purchase due to its high number of positive reviews and strong sales performance. This response case highlights how the interaction mechanism can influence user decision-making again.

\section{Prompts}
\subsection{Subjective Profile Prompt}
The prompt used in Subjective Profile is shown in Table~\ref{tab:subjective_profile_prompt}.
\label{subjective_prompt}
\begin{table*}[htbp]
    \centering
    \begin{tabular}{p{15cm}}
    \toprule
    Please analyze and summarize the user's profile based on the following interaction history between the user and items. Output in JSON format, including the following aspects : \\
    \\
    Taste preferences \\
    Consumption budget range \\
    Price sensitivity \\
    Scenario preferences \\
    Consumption habits \\
    \\
    The output must be in the following JSON format: \\
    \\
    \texttt{\{} \\
    \texttt{  "reason": "Please explain in detail the analysis process and the basis for your conclusions",} \\
    \texttt{  "profile": \{} \\
    \texttt{    "consumption\_budget\_range": "",} \\
    \texttt{    "scenario\_preferences": [],} \\
    \texttt{    "consumption\_habits": [],} \\
    \texttt{    "taste\_preferences": [],} \\
    \texttt{  \}} \\
    \texttt{\}} \\
    \\
    Interaction History: \{input data\} \\
    \bottomrule
    \end{tabular}
     \caption{The prompt used in Subjective Profile.}
    \label{tab:subjective_profile_prompt}
\end{table*}

\subsection{Inferred Profile Prompt}
\label{infer_prompt}
The prompt used in Inferred Profile is shown in Table~\ref{tab:inferred_profile_prompt}.
\begin{table*}[htbp]
    \centering
    \begin{tabular}{p{15cm}}
    \toprule
    Please thoroughly analyze the interaction history between the user and each item, with particular attention to the specific expressions in the user's reviews. From the reviews and items information, mine and summarize the following: \\
    \\
    1. Basic user characteristics: \\
    \hspace*{1em}Estimated age range \\
    \hspace*{1em}Possible occupation type \\
    \hspace*{1em}Estimated income level \\
    \hspace*{1em}Life status (single/married/with children) \\
    \\
    2. Consumption patterns: \\
    \hspace*{1em}Price sensitivity \\
    \hspace*{1em}Quality consciousness \\
    \hspace*{1em}Service preferences \\
    \hspace*{1em}Points of concern \\
    \\
    3. Review language style: \\
    \hspace*{1em}Summarize the user's language style in reviews (e.g., formal, casual, humorous, critical, concise, detailed, emotional, objective, etc.) \\
    \\
    Please form a comprehensive user profile based on this information. When there is insufficient information to make a judgment, please output "unknown" for that item. \\
    \\
    The output must be in the following JSON format: \\
    \\
    \texttt{\{} \\
    \texttt{  "reason": "Please explain in detail the analysis process and the basis for your conclusions",} \\
    \texttt{  "profile": \{} \\
    \texttt{    "estimated\_age\_range": "",} \\
    \texttt{    "possible\_occupation\_type": "",} \\
    \texttt{    "estimated\_income\_level": "",} \\
    \texttt{    "life\_status": "",} \\
    \texttt{    "price\_sensitivity": "",} \\
    \texttt{    "quality\_consciousness": "",} \\
    \texttt{    "service\_preferences": [],} \\
    \texttt{    "points\_of\_concern": [],} \\
    \texttt{    "review\_language\_style": []} \\
    \texttt{  \}} \\
    \texttt{\}} \\
    \\
    Interaction History with Reviews: \{input data\} \\
    \bottomrule
    \end{tabular}
    \caption{The prompt used in Inferred Profile.}
    \label{tab:inferred_profile_prompt}
\end{table*}

\begin{table*}[htbp]
  \centering
    \resizebox{\textwidth}{!}{
    \begin{tabular}{ccccccc}
    \toprule
    \textbf{Pair ID} & \textbf{Evaluator 1} & \textbf{Evaluator 2} & \textbf{Evaluator 3} & \textbf{Human Consensus (Majority)} & \textbf{GPT-4o Judge Agent} & \textbf{Agreement (Human vs. GPT-4o)} \\
    \midrule
    \textbf{1} & A     & A     & A     & A     & A     & Yes \\
    \textbf{2} & B     & A     & B     & B     & B     & Yes \\
    \textbf{3} & A     & B     & A     & A     & A     & Yes \\
    \textbf{4} & B     & B     & B     & B     & B     & Yes \\
    \textbf{5} & A     & A     & B     & A     & A     & Yes \\
    \textbf{6} & B     & B     & A     & B     & B     & Yes \\
    \textbf{7} & A     & A     & A     & A     & A     & Yes \\
    \textbf{8} &       & B     & A     & B     & A     & \textcolor[rgb]{ 1,  0,  0}{No} \\
    \textbf{9} & A     & B     & A     & A     & A     & Yes \\
    \textbf{10} & B     & A     & B     & B     & B     & Yes \\
    \textbf{11} & A     & A     & B     & A     & A     & Yes \\
    \textbf{12} & B     & B     & A     & B     & B     & Yes \\
    \textbf{13} & A     & A     & A     & A     & A     & Yes \\
    \textbf{14} & B     & B     & B     & B     & B     & Yes \\
    \textbf{15} & B     & A     & B     & B     & A     & \textcolor[rgb]{ 1,  0,  0}{No} \\
    \textbf{16} & B     & A     & B     & B     & B     & Yes \\
    \textbf{17} & A     & A     & B     & A     & A     & Yes \\
    \textbf{18} & B     & B     & A     & B     & B     & Yes \\
    \textbf{19} & A     & A     & A     & A     & A     & Yes \\
    \textbf{20} & B     & B     & B     & B     & B     & Yes \\
    \textbf{21} & A     & B     & A     & A     & A     & Yes \\
    \textbf{22} & B     & A     & B     & B     & B     & Yes \\
    \textbf{23} & A     & A     & B     & A     & A     & Yes \\
    \textbf{24} & B     & B     & A     & B     & B     & Yes \\
    \textbf{25} & A     & A     & A     & A     & A     & Yes \\
    \textbf{26} & B     & A     & B     & B     & B     & Yes \\
    \textbf{27} & A     & A     & B     & A     & B     & \textcolor[rgb]{ 1,  0,  0}{No} \\
    \textbf{28} & B     & B     & B     & B     & B     & Yes \\
    \textbf{29} & A     & B     & A     & A     & B     & \textcolor[rgb]{ 1,  0,  0}{No} \\
    \textbf{30} & A     & B     & A     & A     & A     & Yes \\
    \midrule
    \textbf{Total} &       &       &       &       & \textbf{Agreements:} & \textbf{26 / 30 (86.7\%)} \\
    \bottomrule
    \end{tabular}%
    }
    \caption{Detailed human evaluation study results.}
  \label{tab:human_study}%
\end{table*}%

\end{document}